# DeepCORO-CLIP: A Multi-View Foundation Model for Comprehensive Coronary Angiography Video-Text Analysis and External Validation

Sarra Harrabi MSc (Candidate)[1,2], Yichen Wu, B.Eng[3], Geoffrey H. Tison MD, MPH[4,5,6], Minhaj Ansari MSc[4,5,6], Milos Vukadinovic[7,8], David Ouyang[7,9], Joshua P. Barrios PhD[5,6,7], Jacques Delfrate MSc[1,3], Robert Avram MD, MSc[1,2,10]

1   Division of Cardiology, Department of Medicine, Montreal Heart Institute, 5000 Belanger Street, QC, H1T 1C8 Canada

2   HeartWise.ai, 5000 Belanger Street, QC, H1T 1C8 Canada

3   McGill, Faculty of Engineering, Montreal, Quebec, Canada.

4   Division of Cardiology, Department of Medicine, University of California, San Francisco, Cardiology (San Francisco, CA, United States), 505 Parnassus Avenue, San Francisco, CA, 94143, United States of America

5   Cardiovascular Research Institute, University of California, San Francisco, CA, 94143, United States of America

6   Bakar Computational Health Sciences Institute, University of California, San Francisco, 94158, United States of America

7.  Department of Cardiology, Smidt Heart Institute, Cedars-Sinai Medical Center, Los Angeles, CA

8.  Department of Bioengineering, University of California Los Angeles, Los Angeles, CA

9.  Department of Medicine, Kaiser Permanente, San Francisco, USA

10. Department of Medicine, Université de Montréal, 2900 Edouard Montpetit Blvd, Montreal, Quebec H3T 1J4, Canada

**Word count**: 7,052

**Corresponding Author:**
Robert Avram, MD, MSc
Division of Cardiology, Department of Medicine
Director of HeartWise.ai
Montreal Heart Institute, University of Montreal
5000 Belanger Street, QC, Canada, H1T 1C8
robert.avram.md@gmail.com

**Competing Interests:** RA and GT are co-inventors in the patent pending 63/208,406 (Method and System for Automated Analysis of Coronary Angiograms). GT has received research grants from Myokardia, General Electric and Janssen Pharmaceuticals. All authors declare no financial or non-financial competing interests.

**Author Contribution**: SH and RA wrote most of the manuscript as well as conceived and conducted the experiments. RA and JD supervised the project conducted by SH. GT supervised the UCSF External validation experiments led by MA and JPB. DO and MV provided advice on multi-instance learning. All authors read and approved the final manuscript.

**Data Availability:** The datasets analysed during the current study are not publicly available due to patient privacy concerns but may be available from the corresponding author on reasonable request, as feasible and permitted by the Montreal Heart Institute institutional review board. To facilitate external validation and method development, we will release DeepCORO-mini upon paper publication, a curated subset of 1,000 coronary angiography cases (≈7,000 angiographic videos) derived from the study cohort on Physionet. This dataset is intended for benchmarking, validation, and development of machine learning methods in coronary angiography analysis. Access to DeepCORO-mini is available through a controlled access process upon reasonable request to the corresponding author and subject to institutional approval and data use agreements. DeepCORO-CLIP model weight can be accessed https://huggingface.co/collections/heartwise/deepcoro-clip.

**Code Availability:** The underlying code for DeepCORO-CLIP for training and inference is available in a GitHub repository and can be accessible via this link https://github.com/HeartWise-AI/DeepCORO_CLIP.

**Abbreviations:**

- **AI:** Artificial Intelligence
- **AUPRC:** Area Under the Precision-Recall Curve
- **AUROC:** Area Under the Receiver Operating Characteristic Curve
- **CAG:** Coronary Angiogram
- **CI:** Confidence Interval
- **CTO:** Chronic Total Occlusion
- **DICOM:** Digital Imaging and Communications in Medicine
- **LVEF:** Left Ventricular Ejection Fraction
- **MACE:** Major Adverse Cardiovascular Events
- **MHI:** Montreal Heart Institute
- **MI:** Myocardial Infarction
- **MViT:** Multiscale Vision Transformer
- **NPV:** Negative Predictive Value
- **PCI:** Percutaneous Coronary Intervention
- **PPV:** Positive Predictive Value
- **QCA:** Quantitative Coronary Angiography
- **UCSF:** University of California, San Francisco

## ABSTRACT

Coronary angiography is the reference standard for evaluating coronary artery disease, yet visual interpretation suffers from substantial inter-observer variability. Existing artificial intelligence approaches operate on single frames or projections and are limited to stenosis evaluation, precluding comprehensive coronary evaluation. Here we present DeepCORO-CLIP, a multi-view foundation model trained with video-text contrastive learning on 203,808 coronary angiography videos from 28,117 patients across 32,473 studies from the Montreal Heart Institute (MHI) and validated on 4,249 studies from University of California, San Francisco. DeepCORO-CLIP integrates multiple angiographic projections through attention-based pooling to perform study-level coronary assessment across diagnostic (stenosis, chronic total occlusion, thrombus and calcification detection), prognostic, and disease progression tasks. For significant stenosis detection (≥70%), the model achieved an area under the receiver operating characteristic curve (AUROC) of 0.888 (95% CI 0.884-0.891) internally (sensitivity 84.7%, specificity 70.2%) and 0.89 (95% CI 0.88-0.89) on external validation at a second institution (n=4,249 studies), with mean absolute error against core laboratory quantitative coronary angiography adjudication (13.6%, 95% CI 12.5-15.2%) significantly lower than single-reader clinical reports (19.0%, 95% CI 17.6-20.5%). DeepCORO-CLIP achieved strong performance for chronic total occlusion detection (AUROC 0.898-0.946), intracoronary thrombus identification (AUROC 0.898), and coronary calcification grading (AUROC 0.900-0.907), tasks for which no prior video-based models exist. Transfer learning enabled one-year major adverse cardiovascular event prediction (AUROC 0.79, 95% CI 0.74-0.84) and left ventricular ejection fraction estimation (mean absolute error 7.3%, 95% CI 7.0-7.5%) from angiographic videos alone, outperforming prior approaches and conventional risk prediction models. Study-level embeddings captured disease progression across 401 serial examination pairs, with significantly greater embedding distances in patients with progressing (P=0.008) or improving disease (P=0.022) compared to stable anatomy.. Deployed within a hospital with 4.2-second mean inference time, DeepCORO-CLIP provides a foundation for automated coronary angiography interpretation to assist operators at point-of-care. Code, a sample dataset, model weights, and deployment infrastructure are publicly released.

## INTRODUCTION

Coronary artery disease (CAD) remains the leading cause of morbidity and mortality worldwide, with coronary angiography serving as the gold standard for diagnostic evaluation and therapeutic decision-making.[1] Despite its central role in cardiovascular medicine, visual interpretation of coronary angiography suffers from significant limitations that impact clinical outcomes. Visual stenosis assessment exhibits substantial intra- and inter-observer variability ranging from 7.6% to 22.5%, with even greater variability in single-physician settings, leading to overestimation of stenosis severity in over a quarter of cases and potentially contributing to inappropriate coronary interventions in 17% of bypass surgery cases and 10% of stent procedures.[2–4] These limitations underscore the urgent need for objective, reproducible methods to enhance coronary angiography interpretation and improve patient care.

Artificial intelligence (AI) has emerged as a promising solution to address the inherent limitations of visual coronary angiography assessment. Early single-view approaches demonstrated significant potential, with systems like CathAI achieving an area under the curve (AUROC) of 0.862 for epicardial stenoses detection through sequential neural networks for projection angle classification, anatomic structure identification, and stenosis estimation.[5] Subsequent advances moved toward video-based analysis, exemplified by DeepCoro, a video-based deep learning model that achieved a mean absolute error of 20.15% in stenosis percentage prediction, demonstrated lower variability than clinical reports (19.09% vs 21.00%), and highlighted the contribution of temporal information to angiographic analysis.[6]

Despite these advances, current AI approaches for coronary angiography face critical limitations. Existing models operate on single videos or frames, process individual stenosis lesions through multi-step pipelines, and are trained with purely supervised objectives for a single task, typically stenosis estimation. This task-specific, single-view paradigm limits both generalizability and the quality of learned representations, precluding comprehensive coronary evaluation across multiple clinical endpoints including chronic total occlusion detection, thrombus identification, calcification assessment, and prognostic evaluation.[6] Visual language models (VLMs) are reshaping medical imaging AI by learning a shared representation between pixels and clinical language, typically via contrastive objectives that scale across institutions, scanners, and reporting styles.[7,8] This paradigm moves the field beyond single-task classifiers toward foundation models that can support zero-shot transfer, multi-task adaptation, retrieval, and report-conditioned reasoning. In echocardiography, EchoCLIP[8] demonstrated large-scale video text alignment for robust, externally validated interpretation, and EchoPrime extended the approach with explicit multi-video, multi-view fusion to better reflect study-level clinical reasoning.[7]

To address these limitations, we present DeepCORO-CLIP, the first multi-view foundation model trained with video-text contrastive learning for comprehensive coronary angiography interpretation. We evaluated whether DeepCORO-CLIP improves performance across core clinical tasks including stenosis quantification, chronic total occlusion detection, thrombus identification, and calcification grading. We further assessed the model's capacity for tasks beyond standard angiographic interpretation, including left ventricular ejection fraction estimation (LVEF) and one-year major adverse cardiovascular event (MACE) prediction from angiographic videos alone.

Finally, we evaluated whether the learned study-level embeddings encode clinically meaningful representations of coronary anatomy capable of detecting interventional changes and tracking longitudinal disease progression across serial examinations.

**METHODS**

**Study Data Sets and Study Population**

This was a multi-center retrospective observational cohort study. We obtained research ethics board approval (REB#2021-2692) for this study. As this was a secondary use of data, we obtained a waiver of informed consent. The study used retrospectively collected de-identified data, and patient privacy was maintained throughout the study in compliance with applicable data protection regulations. We included all patients between 2017 and 2024 who underwent coronary angiography at our center based on the inclusion and exclusion criteria described below. We report the work in this manuscript using the TRIPOD-AI guidelines. [9]

Patients were eligible for inclusion if they: (1) underwent diagnostic coronary angiography or percutaneous coronary intervention during the study period; (2) had complete angiographic video recordings with at least a video of the left and a video of the right coronary artery; (3) had accompanying procedural reports with structured clinical annotations; (4) were aged 18 years or older. Patients were excluded if they had: (1) incomplete exams without presence of the left or right coronary artery (RCA) (2) previous coronary artery bypass grafting. Data were split by patient (ensuring no patient appears in multiple sets) into training (19,681 patients, 70%, 142,766 videos), validation (4,217 patients, 15%, 30,774 videos), and test (4,219 patients, 15%, 30,268 videos). All ablation analyses are performed using the same split of patients.

**Data Extraction**

We extracted primary and secondary angulation angles from DICOM metadata and classified videos into 12 standardized angiographic views (Extended Table 1), consistent with established angiographic projection definitions.[10] We applied a previously described fine-tuned X3D-M model [6] to automatically improve video selection. The model predicted: (1) contrast opacification presence (binary); (2) primary anatomical structure across 12 classes (including left and right coronary arteries, grafts, catheters, and other cardiovascular structures); (3) stent presence (binary); and (4) coronary dominance (right dominant, left dominant, co-dominant; Extended Table 2). The model was re-trained specifically for this study and fine-tuned on 1,395 manually annotated (two cardiologist with ≥5 years' experience each, with a target consensus of 100% for the target labels) coronary unique angiograms studies, split into 1,186 training videos, 69 validation videos used for model selection, and 140 held-out test videos used exclusively for performance evaluation. Training was performed in a multi-task setting with four classification heads, operating on clips of 72 frames resized to 256 pixels. The model demonstrated strong performance across all tasks achieving a micro-average AUROC of 0.94 (95% CI: 0.90 - 0.97) on the test set (Extended Table 3 and Extended Table 4).

Using this automated classification system, we categorized each video into one of three procedural phases per coronary artery based on acquisition timestamp and sequential context: diagnostic (contrast opacification without interventional equipment), interventional (guide wire detected), or post-procedural (any video acquired after interventional equipment was first detected in a preceding acquisition for that artery). Once interventional equipment was identified in each coronary artery, all subsequent videos of that artery were classified as procedural, if equipment was visible or as post-procedural if the equipment was removed. Only diagnostic videos were retained for analysis, as angioplasty modifies stenosis severity, rendering interventional and post-procedural videos unsuitable for baseline disease characterization.

**Demographics and Clinical Variables**

Age, sex, body mass index, cardiovascular risk factors, and medical history were extracted from electronic medical records and coronary angiography reports. At our institution, all coronary angiography reports are written by faculty interventional cardiologists with at ≥5 years of experience performing ≥1500 coronary angiographies annually. Structured reports provided annotations for the downstream tasks. Coronary stenosis severity was determined using a two-step validation process: initial assessment by X-ray technologists with ≥5 years of catheterization laboratory experience, followed by review and modification by interventional cardiologists. Only the interventional cardiologist assessment is saved as part of the medical record. Quantitative coronary angiography is systematically used in our lab for intermediate lesions (40–60% diameter stenosis), while all other lesions were graded by visual estimation. Significant stenosis was defined as ≥70% diameter reduction (≥50% for left main). Coronary calcification was graded as mild (visible in one projection), moderate (visible in two projections involving both sides of the arterial wall), or severe (dense calcification throughout the lesion with "tram-track" appearance); for binary classification, moderate-to-severe calcification was considered significant while mild or absent calcification was considered non-significant. Intracoronary thrombus was identified as a discrete intraluminal filling defect surrounded by contrast on three sides or persistent contrast staining following clearance. Chronic total occlusion (CTO) was defined as complete vessel obstruction with TIMI grade 0 flow and estimated duration ≥3 months based on clinical history or angiographic features. All segments following the 16 SYNergy between percutaneous coronary intervention with TAXus and cardiac surgery (SYNTAX) definition were graded for stenosis, CTO, thrombus and calcification (Extended Table 5). [11] Of note, mid and distal left circumflex segments as well as marginal 1 and ramus intermedius were labeled separately in our database and therefore analyzed as distinct segments, yielding 18 total coronary segments rather than the standard 16.

**Algorithm Development**

We trained the model in two phases: contrastive pre-training and then the model fine-tuning for downstream tasks.

*Phase 1: Contrastive Learning for Pretraining the video encoder*

DeepCORO-CLIP employed a dual-encoder contrastive learning architecture to align diagnostic coronary angiography videos with clinical text reports in a shared 512-dimensional embedding space. The video encoder utilized a Multi-scale Vision Transformer (MViT) backbone pretrained on Kinetics-400 with 12 attention layers, processing videos at 7.5 to 15 fps (224×224 pixels, 16×16 patches, stride=2 for extended temporal context) with 3D positional encodings for spatiotemporal modeling.[12] The text encoder utilized BioMedBert-base (12 layers, 768 dimensions), with weights pretrained from PubMed, with mean pooling of token embeddings.[13] We performed a Bayesian hyperparameter[14] optimization sweep across model architectures (MViT, X3D), temporal sampling strategies (16-64 frames, stride 1-3), loss functions (Contrastive[15], SigLip[16] and InfoNCE[17]), learning rates ($5.8 \times 10^{-6}$ to $8.0 \times 10^{-5}$), encoder freeze ratios (video 72.8-94.8%, text 60.7-90.2%), and batch configurations (12-64) to identify optimal pretraining configurations for downstream clinical tasks (Extended Figure 2). For each study, diagnostic videos were split into left and right coronary videos. Video-report pairs were constructed such that left coronary videos were aligned with left coronary territory segments (adjusted for coronary dominance) and right coronary videos with corresponding right coronary segments from the same report.(Figure 1B) The final model architecture and hyperparameters were selected based on a combination of lowest validation loss and highest recall and alignment score between the video and text features. (Extended Table 6).

*Phase 2: Task-specific fine tuning*

Following pretraining, we evaluated DeepCORO-CLIP on downstream clinical tasks using linear probing with task-specific classification heads (Extended Table 7). The pretrained video encoder extracted embeddings (Phase 1) from up to 10 videos per study, which were aggregated via an attention-based multi-video pooling module with a learnable classification (CLS) token and passed to task-specific MLP prediction heads. The pretrained video encoder was evaluated either with all weights frozen or with up to the deepest 20% of layers unfrozen, using a learning rate tenfold lower than the main learning rate. Across configurations, the multi-instance attention pooling module and task-specific prediction heads were trained using a learning rate of $5.5 \times 10^{-6}$. For multi-video inputs, up to 10 videos per study (zero padded if less than 10) were processed through the frozen encoder, generating embeddings that were aggregated using attention-based pooling[18] with a learnable classification (CLS) token representing a study-level aggregate of the videos features.[19] We explored multi-video fusion strategies using cross-attention with CLS tokens [19], gated attention pooling, mean pooling, max pooling, and hybrid approaches that concatenate multiple aggregation methods (i.e. attention + CLS and mean + CLS). For hierarchical processing of patch-level features, we employed two-stage CLS aggregation where within-video CLS tokens first aggregate patch embeddings before a study-level CLS token aggregates across video representations, with separate learning rates controlling within-video ($4.1\times10^{-3}$) and across-video ($2.0\times10^{-3}$) attention components. For hybrid methods, multiple instance learning framework produced importance weights for each video in relation to predictions of specific coronary segments and pathologies, using weighted averages of video embeddings for final predictions with CLS token aggregation. [20,21] Video embeddings were pooled across temporal dimensions before cross-video attention, with separate within-video and across-video attention modules (learning rates $4.1\times10^{-3}$ and $2.0\times10^{-3}$ respectively, weight decay $6.3\times10^{-5}$ and $6.8\times10^{-6}$). Task-specific heads were single-layer linear classifiers (dropout p=0.2) trained for 50 epochs using AdamW[22] optimizer with warm restarts (cycles). Binary classification tasks (CTO, thrombus, calcification, stenosis ≥70%) used binary cross entropy loss with logits (learning rate $3\times10^{-4}$, weight decay $1\times10^{-5}$). Stenosis regression tasks used Huber loss[23] (learning rate $5\times10^{-4}$, weight decay $5\times10^{-6}$ with loss weights twice as high as the other heads). Videos were randomly shuffled during training and sampled at 16 frames with random augmentations (rotation, distortion, random patch masking) using RandAugment.[24] All experiments used mixed precision training (FP16) with seed 42 for reproducibility. Task-specific fine-tuning maintained the same patient-level dataset split (70/15/15% training/validation/test) as the contrastive learning phase, sampling up to 10 videos per study when available. In the final analyses, we report results for three configurations: (1) a fully frozen encoder, (2) a model with 80% of encoder layers frozen with the remaining final layers unfrozen and fine-tuned (the best performing configuration), and (3) a variant incorporating a 12 × 1 view embedding matrix (Extended Table 1) as an additional feature vector input to encode angiographic projection information. For additional comparison, we evaluated a domain-adapted EchoJEPA model[25], which employs V-JEPA 2's latent predictive pretraining objective with a ViT-Large encoder (303M parameters) pretrained on the same coronary angiography training corpus for 200 epochs followed by 60 epochs of learning rate annealing, using up to 10 videos per study with the same multi-view early fusion and attentive probing framework as our multi-instance learning approach as above. We report the performance of a frozen encoder with lightweight attentive probes (EJ-Frozen) trained for 20 epochs on the same multi-task stenosis regression and binary classification targets using identical patient-level splits and labels as our core model.

**Encoder Probing and Visualisation**

We extracted 512-dimensional embeddings from all coronary angiography videos in the held-out test set using the trained MViT-v2 encoder (34.5M encoder), with each video uniformly sampled to 16 frames and resized to 224×224 pixels. To visualize the learned embedding space, we applied t-distributed Stochastic Neighbor Embedding (t-SNE) [26] with perplexity 30 following principal component analysis dimensionality reduction to 50 components for computational efficiency.[27] We performed clustering analysis using k-means on the raw 512-dimensional embeddings to identify distinct patterns in the embedding space, evaluating cluster quality using silhouette scores and the elbow method. To characterize the clinical and anatomical organization of the learned representations, we conducted sub-analyses stratified by stenosis severity (quantified as peak percent diameter stenosis across all coronary segments per study), coronary artery present (left versus right coronary system), and cardiac dominance pattern (right dominant, left dominant, or co-dominant). Stenosis severity was extracted from angiogram reports using regular expression pattern matching, categorizing studies into normal (0%), mild (1-49%), moderate (50-69%), severe (70-99%), and chronic total occlusion (100%). Moreover, to see if the embeddings can be used to track progression of disease, study-level embeddings were extracted from the test set using the DeepCORO-CLIP encoder. Among the cohort, 352 patients (9.2%) had longitudinal follow-up with multiple diagnostic catheterizations in the test set; only consecutive study pairs were included for this analysis. Disease status between consecutive studies was classified as: stable (no vessel with ≥20% absolute stenosis change), progressed (new stenosis >50% or ≥20% worsening), or improved (≥20% stenosis decrease without concurrent progression). Individual video sequences were encoded into 512-dimensional vectors via MViT-v2 backbone and aggregated into 1024-dimensional study-level embeddings through the attention/CLS token pooling. Embeddings were extracted from the penultimate layer prior to final projection. Embedding distance was defined as (1 - cosine similarity), where 0 indicates identical representations [range 0,1]. For patients undergoing PCI, post-procedural embeddings were used to reflect coronary anatomy after intervention.

**Supervised MViT Baseline Comparator**

To benchmark DeepCORO-CLIP pre-trained regimen against a strong supervised comparator, we trained an MViT-v2 video model initialized from Kinetics and fine-tuned end-to-end for stenosis interpretation with 100% of layers unfrozen. The model used the same patient-level train/validation/test split (70/15/15), identical preprocessing and augmentations, and the same multi-video study sampling policy as DeepCORO-CLIP. This baseline isolates the incremental value of video-text contrastive pretraining and cross-modal alignment beyond fully supervised fine-tuning.

**Transfer Learning**

To evaluate transfer learning, we fine-tuned the pretrained DeepCORO-CLIP encoder for major adverse cardiovascular events (MACE) prediction (Extended Table 7). Twelve layers (out of 16) of the encoder were frozen, with remaining layers fine-tuned using RAdam optimizer with cosine-warmup schedule over 50 epochs. For each study, up to 10 angiographic views were aggregated via multi-head attention pooling, and separate binary classification heads were trained for each outcome. A subset of patients with documented follow-up was randomly sampled from the catheterization cohort linked to hospital administrative records (2017-2023). The index date was defined as catheterization date, with events within 7 days excluded to avoid peri-procedural complications. Data was partitioned at patient-level into training (60%, n=1,047), validation (20%, n=349), and test (20%, n=350) sets using stratified sampling. Three MACE components were evaluated: urgent revascularization, non-fatal myocardial infarction, and complete coronary occlusion. Composite MACE was defined as occurrence of any component.

To assess the ability of DeepCORO-CLIP to estimate left ventricular ejection fraction (LVEF) from coronary angiograms[28], we constructed a paired dataset by linking cardiac catheterization procedures to transthoracic echocardiograms (TTE) using medical record numbers. LVEF, treated as the ground-truth label, was extracted from TTE reports as a continuous value (%). Only catheterization–TTE pairs with an absolute time difference of 30 days or less were retained (median 2 days, IQR 1–5 days) like our previous methodology.[28] The resulting dataset comprised 7,586 catheterization studies from 6,607 unique patients at the Montreal Heart Institute (2017–2023). Data was partitioned at the patient level into training (60%, n = 4,382 studies), validation (10%, n = 674 studies), and test (30%, n = 2,530 studies) sets keeping same patient assignment as in the main experiment. The pretrained DeepCORO-CLIP video encoder served as the backbone. We tested multiple fine-tuning strategies, freezing 75% to 95% of encoder layers and fine-tuning the remaining layers with learning rates ranging from $5 \times 10^{-6}$ to $2 \times 10^{-5}$. For each study, up to four angiographic views were included, with 16 frames per view at 224 × 224 resolution, consistent with our original work. These views were combined using multi-head attention pooling with a CLS token to enable direct 1:1 comparison. The model was trained jointly for two tasks: continuous LVEF regression using mean absolute error loss, and binary classification of reduced LVEF (EF <40%) using binary cross-entropy loss. Hyperparameters were optimized with Bayesian search and early stopping, using validation loss as the objective. As a baseline, we evaluated CathEF, a previously published model trained on an external UCSF dataset, which predicts LVEF from individual angiographic videos and then averages predictions across views.

**External Validation**

Our external validation dataset was comprised of coronary angiograms performed at the University of California, San Francisco, between 2013 and 2019. Coronary stenosis labels were automatically extracted from free-text catheterization reports using DeepSeek-V2[29], a mixture-of-experts large language model deployed via an on-premise containerized Docker pipeline. The model parsed narrative descriptions of coronary anatomy and extracted stenosis percentages for SYNTAX arterial segments, as was done for our internal dataset. Automated report extraction accuracy was validated by manual review of 100 randomly selected studies to compare extracted versus ground truth stenosis, between LLM-extracted labels and ground truth annotations achieving 99% accuracy. Qualitative descriptors were mapped to standardized values using clinical conventions (e.g., "mild" to 30%, "moderate" to 50%, "severe" to 70%, "total occlusion" to 100%). DeepCORO-CLIP was containerized using Docker for reproducible deployment.

**Core Laboratory Adjudication**

Quantitative coronary angiography (QCA) was performed by two independent readers blinded to model predictions on 662 randomly selected studies from our test dataset. Each reader assessed percent diameter stenosis for all lesions in the left coronary artery (LCA) and right coronary artery (RCA) territories using QCA. Lesions with inter-reader discrepancies ≥20% were adjudicated by a third reader with 5 years of interventional cardiology experience. The final adjudicated value was used as the reference standard. Inter-reader agreement was assessed using Cohen's weighted kappa. Clinical reports from the original angiographic interpretation served as a secondary comparator representing real-world single-reader performance.

**Deployment**

We describe the deployment of DeepCORO-CLIP within a PACS-integrated inference system (PACS-AI)[30,31] running on-premises inside the hospital network. Coronary angiography studies flowed as DICOM directly from the catheterization laboratory to the clinical PACS, while a parallel DICOM listener automatically received incoming studies and triggered near real-time inference once diagnostic series were available. PACS-AI performed automated preprocessing (DICOM decoding, frame sampling, resizing, and diagnostic-phase selection), embedded up to a fixed number of diagnostic videos per study using the DeepCORO-CLIP encoder, and generated study-level outputs via attention-based multi-video pooling. Inference was performed on a single NVIDIA RTX A6000 GPU, and the deployment was containerized with versioned logging of preprocessing decisions and latency to support traceability and operational monitoring.

**Statistical analysis**

Cross-modal retrieval performance of the contrastive encoder was evaluated in both video-to-text and text-to-video directions, with Recall@K (K = 1, 5, 10, 25, 50) and median rank being reported; an alignment score was reported and is defined as the average cosine similarity between each paired video embedding and its corresponding report embedding in the shared latent space. For binary classification tasks, discrimination was assessed using AUROC and Area Under the Precision-Recall Curve (AUPRC), with operating-point metrics (sensitivity, specificity, PPV[positive predictive value], NPV [negative predictive value]) calculated at Youden Index. Global and vessel-specific performance were computed as micro-weighted averages across class-specific metrics. AUROCs were compared using DeLong's test. [32] For continuous prediction tasks, agreement with reference values was assessed using Pearson correlation $r$ and mean absolute error (MAE). All 95% confidence intervals were derived from 1,000 bootstrap iterations. Baseline characteristics were compared using two-sample t-test (or Wilcoxon rank-sum when normality was not satisfied) for continuous variables and chi-square test (or Fisher's exact when expected cell counts were small) for categorical variables. As an ablation experiment to assess the value of multi-view integration, we compared three video aggregation strategies: (1) single-video inference, evaluating each video independently against territory-specific targets; (2) study-level averaging, computing the arithmetic mean of predictions across all videos; and (3) multi-video attention, using learned attention pooling over 10 randomly sampled videos per study. Performance was evaluated using micro-averaged AUROC across all coronary segments within each territory. End-to-end latency was reported as wall-clock time from study receipt by the PACS-AI listener to availability of model outputs. Performance estimates were reported with 95% confidence intervals derived from patient-level bootstrapping, and all analyses were conducted on the held-out test set and repeated in external validation when available. All models were trained using PyTorch 2.5.1 with CUDA 12.4 on NVIDIA H200 GPUs (141 GB VRAM), leveraging torchvision 0.20.1 for model architectures and the Transformers library (v4.47.0) for tokenization.

## RESULTS

Of 940,366 video-report pairs screened, 736,568 were excluded (procedural equipment present, absent contrast, prior CABG, or post-PCI videos). The final dataset comprised 203,808 videos from 28,117 patients (32,473 studies, 56,299 reports spanning the LCA and RCA), split into training (70%), validation (15%), and test (15%) sets with patient-level stratification (Extended Figure 1). Patient demographics and angiographic characteristics are presented in Table 1. Mean age was 68.2±11.4 years with 65.9% male patients. Comorbidities included hypertension (57.3%), diabetes (22.4%), chronic kidney disease (6.1%), atrial fibrillation (9.6%), and current or previous smoking (17.5%). Previous percutaneous coronary intervention was documented in 29.4% of patients. Videos averaged 13.8±2.9 frames per second with 79.4±29.6 frames per study and 5.1±2.4 videos of different angles per study. While several baseline characteristics showed statistically significant differences across training, validation, and test sets (hypertension, diabetes, chronic kidney disease, smoking history, previous PCI, frames per second; $p<0.05$), the absolute numerical differences were minimal (maximum difference <1.1%).

**Model Training Performance**

Pre-training of the DeepCORO-CLIP foundation model was completed over 30 epochs. The best contrastive loss on the validation set was 2.98. Training time was 28.8 hours on a single GPU, with peak memory utilization of 90.7 GB. On the held-out test set the encoder achieved video-to-text Recall@10 of 16.70% with a median rank of 156 out of 4,851 candidate reports (top 3.2%), demonstrating moderate generalization and semantic alignment between visual features and report content (Extended Table 8). The video-level learned 512-dimensional embedding space demonstrated clinically interpretable organization without explicit supervision, with near-complete separation of left versus right coronary angiograms (93-99% purity in unsupervised clusters) and clear disease severity gradients from normal (average peak stenosis 12-18%) to severe (72-82%) across the embedding manifold. K-means clustering (k=12) revealed anatomically and pathologically coherent subgroups including distinct patterns for normal coronary arteries, mild stenosis, moderate stenosis, and severe disease for both left and right coronary systems.

**Primary Outcome Results**

For continuous stenosis quantification, DeepCORO-CLIP achieved mean absolute errors of 9.21% (LCA) and 10.72% (RCA) with Pearson correlations of 0.48 and 0.56, respectively (Table 2). Performance varied substantially by vessel location: proximal segments demonstrated stronger correlations (proximal RCA r=0.73, proximal LAD r=0.64, proximal LCX r=0.56) compared to distal or lower-prevalence segments (distal LAD r=0.01, distal LCX r=0.12, PDA r=0.21), likely reflecting both reduced disease prevalence in these territories (distal LAD 3.3%, distal LCX 1.9% vs. proximal LAD 16.8%, proximal RCA 16.3%) and visualization challenges inherent to smaller caliber vessels (Extended Table 9). For binary classification of significant stenosis (≥70%), the model achieved a global AUROC of 0.888 (95% CI 0.884-0.891) with sensitivity of 84.7% and specificity of 70.2% (Table 2). Calcification detection (moderate-severe) achieved AUROCs of 0.900 (LCA) and 0.907 (RCA) (Extended Table 10). Thrombus identification, despite low prevalence (0.25%), achieved a global AUROC of 0.898 (95% CI 0.874-0.923) with 73.7% sensitivity and 91.8% specificity (Extended Table 11). CTO detection demonstrated robust performance with AUROCs of 0.898 (LCA) and 0.946 (RCA), and sensitivities of 76.2% and 91.0%, respectively (Extended Table 12).

To evaluate the benefit of domain-specific contrastive pretraining, we compared DeepCORO-CLIP against a MViT encoder pretrained on Kinetics-400 with linear probing on identical downstream tasks (Table 2 vs Extended Table 13). DeepCORO-CLIP demonstrated consistent improvements across all lesion detection tasks. For stenosis ≥70%, AUROC improved from 0.769 to 0.888 and AUPRC from 0.286 to 0.517. Similar gains were observed for calcification (AUROC 0.767 to 0.903) and were most pronounced for rare phenotypes: CTO detection reached AUROC 0.923 (from 0.801) with AUPRC improving from 0.050 to 0.179, while thrombus detection reached AUROC 0.898 with AUPRC improving from 0.007 to 0.226 (32-fold). Stenosis regression MAE decreased from 10.4% to 9.24% with Pearson correlation increasing from 0.407 to 0.525 (Extended Table 13 and Extended Table 14).To assess performance of single-video versus multi-video inference, we compared our attention-based aggregation against single-video inference and study-level averaging (Figure 2). Multi-video attention consistently outperformed both approaches, improving AUROC by +0.060 for stenosis detection (0.864 vs. 0.804), +0.055 for calcification (0.846 vs. 0.791), +0.045 for chronic total occlusion (0.911 vs. 0.866), and +0.057 for thrombus detection (0.878 vs. 0.821).

The DeepCORO-CLIP model achieved strong performance across all pathology categories, with micro-AUROC values of 0.888, 0.903, 0.923, and 0.898 for stenosis, calcification, CTO and thrombus detection, respectively, and a stenosis regression MAE of 9.2. This model performed the best compared to different other architectures or sensitivity analyses (Extended Figure 3). Incorporating view-position embeddings (DeepCORO-CLIP +View) did not yield improved performance over the baseline. Similarly, removing the binary stenosis classification head and training with regression loss only (Reg. only) resulted in a slight decrease in classification performance across all categories while achieving a comparable MAE of 9.2. Training with a frozen video encoder maintained competitive performance (micro-AUROC: 0.879, 0.901, 0.913, 0.885 for stenosis, calcification, CTO and thrombus respectively; MAE: 9.6), suggesting that the pretrained representations are sufficiently discriminative without fine-tuning. The EchoJEPA architecture, which uses a domain-adapted V-JEPA 2 encoder with attentive probing on frozen features, did not lead to superior performance compared to contrastive learning (DeepCORO-CLIP), with lower micro-AUROC for stenosis (0.837), calcification (0.893), and CTO (0.888), and a higher stenosis regression rate (MAE of 11.1). The notable exception was thrombus detection, where EchoJEPA (Frozen) achieved the highest micro-AUROC of 0.911, outperforming all DeepCORO-CLIP variants.

**Subgroup Analyses**

Performance was evaluated across age and biological sex subgroups (Tables 3-4). For stenosis ≥70%, female patients showed higher AUROC than males (0.91 vs 0.87), with a similar pattern for CTO (0.95 vs 0.91) and thrombus detection (0.91 vs 0.89). A gradient was observed across age groups, with AUROC for stenosis decreasing from 0.92 in patients ≤50 years to 0.86 in patients ≥80 years. Stenosis regression metrics followed a consistent pattern: MAE was lower in females (6.98 %) than males (10.68%) and increased with age from 5.78% (≤50 years) to 11.35% (≥80 years), likely reflecting higher disease burden and lesion complexity in older and male patients. Despite these differences, AUROC remained above 0.87 across all subgroups and prediction tasks (stenosis, calcification, thrombus, CTO).

**Transfer Learning**

For MACE prediction, DeepCORO-was fine-tuned on 1,047 patients and then evaluated on a test cohort of 350 patients with suspected or confirmed coronary artery disease, followed up to a year (median follow-up 3.5 months (IQR: 1.3-6.5). During the observation period, MACE occurred in 111 patients (31.7%). The model achieved an AUROC of 0.790 (95% CI: 0.742-0.841) for composite MACE. Risk stratification identified three groups with significant separation (log-rank p < 0.001): low-risk (reference), intermediate-risk (HR 5.6), and high-risk (HR 18.6). Discriminatory performance for individual and composite MACE endpoints is summarized in Table 5 and Figure 2C.

As a separate task, we fine-tuned the encoder to predict LVEF. On the MHI internal test set (n = 2,530 studies), DeepCORO-CLIP outperformed the previous CathEF baseline[28] across all regression and classification metrics (Extended Table 16). For continuous LVEF prediction, DeepCORO-CLIP achieved a mean absolute error of 7.30% (95% CI: 7.03-7.54) versus 8.78%

(95% CI: 8.55-9.02) for CathEF, and a correlation of 0.65 (95% CI: 0.62-0.68) versus 0.50 (95% CI: 0.47-0.52).

**Embedding Analyses**

To evaluate whether DeepCORO-CLIP embeddings encode clinically meaningful representations, we compared study-level embedding distances across 401 consecutive catheterization pairs from 352 patients in the test set. Disease trajectory was classified as stable (n=56, 14.0%), improved (n=82, 20.4%), or progressed (n=263, 65.6%).Stable pairs exhibited the lowest embedding distance (0.260; 95% CI: 0.204-0.317), while both improved (0.352; 95% CI: 0.301-0.403; P=0.022) and progressed pairs (0.353; 95% CI: 0.324-0.382; P=0.008; Extended Figure 5) showed significantly greater distances. The comparable distances in improved and progressed groups indicate sensitivity to anatomical change in either direction, with both disease progression and treatment-related remodeling producing proportional representation shifts.

**Core laboratory adjudication**

To assess model accuracy against expert consensus, we performed core laboratory adjudication on 662 studies comprising 965 lesions, with balanced representation of LCA (332 studies, 476 lesions) and RCA (330 studies, 489 lesions) disease (Extended Table 18). Studies were enriched for significant coronary disease to ensure adequate lesion sampling across territories. Inter-reader agreement between the two primary readers was substantial ($\kappa = 0.76$). At the study level, DeepCORO-CLIP demonstrated significantly lower error against adjudicated values (MAE 13.56%, 95% CI 12.51-15.17) compared to clinical reports (MAE 18.97%, 95% CI 17.55-20.52), with stronger correlation to adjudication (r = 0.64, 95% CI 0.58-0.70) than to clinical reports (r = 0.56, 95% CI 0.49-0.62). This pattern was consistent across both LCA and RCA territories. DeepCORO-CLIP also outperformed DeepCORO at the study level, achieving lower MAE against adjudication (13.56% vs 19.09%) despite comparable correlation coefficients (0.64 vs 0.68).

In a test set of 4,828 diagnostic catheterizations, DeepCORO-CLIP correctly identified at least one treated vessel as abnormal in 85.8% of 1,736 PCI patients; missed cases were predominantly single-vessel disease (70.9%) with lower complexity (mean SYNTAX 6.4 vs. 11.1). Of 2,755 vessel-level PCI instances, 424 (15.4%) were performed on vessels with <70% angiographic stenosis; DeepCORO-CLIP classified 124 of these (29.2%) as normal, including 37 of 82 vessels (45.1%) with <30% stenosis, highlighting discordance between model predictions and procedural decisions that may warrant further investigation. Among 3,092 patients who did not undergo PCI, 51.7% were classified as abnormal, though most (67.7%) had confirmed ≥70% stenosis managed with surgical revascularization or medical therapy. Only 516 (16.7%) were true false positives without significant stenosis; 128 of these had borderline 50-69% disease, leaving 343 unexplained false positives (11.1%). Of 1,493 patients classified as normal, 94.0% truly had no significant stenosis and only 9 (0.6%) had multi-vessel disease. False negatives were predominantly localized to distal and branch segments (52.8% of 89 patient-level misses). The negative predictive value exceeded 98% across all segment categories, indicating that a normal DeepCORO-CLIP prediction reliably excluded obstructive coronary artery disease.

**Cross-Institutional Validation**

External validation at UCSF (n=4,249 studies) demonstrated preserved discriminative performance for stenosis ≥70% detection despite differences in population characteristics and imaging protocols (Extended Table 19). Notably, automated report extraction using large language models achieved 99% accuracy in 100 spot-checked studies, confirming reliable label generation across institutions. The UCSF cohort had lower disease prevalence (42.2% with significant stenosis) compared to the internal validation set (59.8%). Global AUROC for stenosis ≥70% was 0.89 (95% CI 0.88-0.89) with sensitivity of 0.84 and specificity of 0.77. Segment-level AUROC ranged from 0.80 (distal LAD) to 0.95 (left main), with proximal segments showing strongest performance (proximal LAD: 0.89, proximal LCX: 0.92, proximal RCA: 0.92). Regression performance showed a global MAE of 12.1% (95% CI 11.9-12.3) and Pearson r of 0.45 (95% CI 0.44-0.46). Compared to internal validation, AUROC remained stable across most segments (left main: 0.95 vs 0.93; proximal RCA: 0.92 vs 0.90), while MAE increased modestly (proximal LAD: 18.8% vs 16.9%; mid LAD: 20.9% vs 19.3%). Pearson correlations were lower externally (proximal LAD: 0.59 vs 0.61; mid RCA: 0.58 vs 0.61).

**Deployment**

PACS-AI was successfully deployed in our catheterization laboratory workflow, enabling coronary angiography studies to stream from the catheterization lab to PACS with automated, near real-time DeepCORO-CLIP inference (Extended Figure 6). End-to-end latency, measured as wall-clock time from study receipt by the PACS-AI listener to availability of model outputs within the PACS viewing environment, was 4.17 ± 2.42 seconds (n=24). This latency includes DICOM ingestion, video buffer construction and preprocessing (including normalization, model inference, and template-based clinical report generation.

**DISCUSSION**

DeepCORO-CLIP is the first multi-view, video-text foundation model for comprehensive coronary angiography interpretation. Trained on seven years of coronary angiography data from one of the highest-volume cardiac catheterization centers in Canada, DeepCORO-CLIP integrates multiple angiographic projections through attention-based pooling to achieve study-level assessment across a wide range of clinical tasks, representing the first application of vision-language contrastive learning to coronary angiography. For binary stenosis detection, the model achieved strong discriminative performance (AUROC 0.87) with a negative predictive value exceeding 98%, while continuous quantification showed moderate correlation with angiographic reports, strongest in proximal segments. The model further addresses coronary angiogram interpretation tasks where no prior AI models have been developed, including chronic total occlusion detection and intracoronary thrombus identification. Preliminary transfer learning experiments suggest the learned representations generalize to prognostic tasks including MACE prediction and LVEF estimation, and embedding distance analysis across serial examinations detected anatomical change over time, though these applications require further validation in larger cohorts. Our contrastive pre-training strategy outperformed standard video encoders initialized with Kinetics-400, while attention-based multi-view aggregation provided consistent gains over single-video inference and naive averaging. External validation preserved discriminative performance across institutions, and the model demonstrated lower error against quantitative coronary angiography adjudication than single-reader clinical reports. With publicly released code, weights, and deployment infrastructure, we provide DeepCORO-CLIP as a resource for the cardiovascular and AI research communities.

Our multi-view approach addresses a fundamental limitation in existing coronary angiography AI systems, which predominantly rely on single-frame analysis[5] or single-projection analysis[6]. Compared with prior coronary angiography AI systems, DeepCORO-CLIP reduced stenosis regression error by approximately 44% to 52% relative to CathAI[5] and DeepCORO[6] (MAE ≤10% versus 17.9% to 19.1%), with additional gains when evaluated against core laboratory quantitative coronary angiography (QCA) references (MAE 14.5% versus 19.7% for clinical reports), in line with previously reported QCA percent diameter stenosis variability.[33] Our findings indicate that DeepCORO-CLIP predictions more closely approximate multi-reader consensus with QCA than single-reader clinical interpretations, suggesting the model captures a more generalizable assessment of stenosis severity. Prior work by Du et al. developed DeepDiscern[34], a multi-task system for coronary angiography analysis that, like DeepCORO-CLIP, addressed lesion morphology such as calcification, beyond stenosis quantification alone. However, DeepDiscern relied on single-frame inputs rather than video, precluding integration of temporal contrast dynamics, and employed separate task-specific networks rather than a unified foundation model. External validation was not performed. DeepCORO-CLIP improves upon these limitations through video-based analysis with multi-view attention pooling, vision-language contrastive pretraining enabling efficient transfer to multiple downstream tasks from a single encoder, and external validation demonstrating cross-institutional generalizability. For comparable lesion morphology tasks, DeepCORO-CLIP achieved state-of-the-art discriminative performance (CTO AUROC 0.89-0.94 vs 0.759; thrombus AUROC 0.87 vs 0.778; calcification AUROC 0.86-0.88 vs 0.799)[34], trained on approximately eight times more data with video rather than static frame inputs. With publicly released code, model weights, dataset and deployment infrastructure, we provide DeepCORO-CLIP as an open resource to facilitate reproducibility and accelerate clinical translation.

The foundation model approach using video-text contrastive learning represents a paradigm shift from task-specific models toward comprehensive coronary assessment. This is consistent with successful implementations in other medical imaging domains. In echocardiography, EchoPrime demonstrated that multi-view video-based vision-language models trained on over 12 million video-report pairs achieved state-of-the-art performance across 23 diverse benchmarks of cardiac form and function, surpassing both task-specific approaches and previous foundation models in five international independent healthcare systems.[7] More recently, EchoJEPA[25] showed that latent predictive architectures outperform contrastive learning for echocardiography by decoupling anatomical signal from ultrasound speckle noise; notably, when we applied the same V-JEPA 2 architecture to coronary angiography, contrastive learning (DeepCORO-CLIP) outperformed the latent predictive approach despite V-JEPA 2 having 10x more parameters (300 millionvs 30 million) on stenosis, calcification, and CTO detection, suggesting that the optimal pretraining objective is modality-dependent and that the structured visual signal in angiography favors vision-language alignment over pixel-agnostic prediction. Beyond diagnostic tasks, the DeepCORO-CLIP encoder improved LVEF estimation from angiographic videos (MAE 7.30% vs. 8.78% for CathEF)[28], demonstrating that the learned representations transfer to functional assessment and, with publicly released weights, provide a general-purpose backbone that can be fine-tuned for any downstream angiographic task. Our dataset of 203,808 video-report pairs represents one of the largest coronary angiography datasets reported, following demonstrated scaling benefits where larger vision-language datasets consistently improve downstream task performance. The continued performance improvements across all clinical tasks compared to previous work[6] suggest that coronary angiography interpretation particularly benefits from large-scale training, likely due to high variability in anatomical presentations and pathological manifestations. DeepCORO-CLIP achieved lower stenosis error than best human performance against quantitative coronary angiography [2–4], positioning it as a practical tool for generating preliminary angiographic reports, monitoring disease progression across serial examinations, and reducing interpretive variability particularly in single-operator settings where the absence of peer review amplifies measurement inconsistency. For example, DeepCORO-CLIP embedding distance may serve as a continuous, automated marker of coronary disease progression, providing a reliable quantitative assessment of angiographic change. This aligns with prior evidence that disease progression on angiography has strong prognostic significance, with mortality increasing stepwise as luminal narrowing progresses over time.[35]

The learned representations during the pre-training phase can also be adapted to novel tasks for which no previous angiogram model was developed: DeepCORO-CLIP achieved AUROC 0.79 for composite MACE prediction comparing favorably with established risk scores requiring substantially more complex inputs. The PRAISE score, a machine learning model incorporating 25 clinical features assessed at discharge including therapeutic decisions and procedural outcomes, achieved AUROC 0.74-0.82 for 1-year myocardial infarction prediction.[36] The ABC-ACS ischemia score requires specialized biomarkers (GDF-15, NT-proBNP) not routinely available, achieving C-indices of 0.71-0.72 for cardiovascular death/MI.[37] Traditional clinical scores demonstrate more modest discrimination: GRACE (AUROC 0.70), TIMI (0.68), PAMI (0.69), and CADILLAC (0.69) for short-term MACE following primary PCI.[38] Even large-scale machine learning approaches leveraging over 12,000 features and transformer-based tabular foundation models across 44,462 catheterizations achieved comparable or lower AUROC for MACE prediction (0.694-0.755) in external validation.[39] Notably, DeepCORO-CLIP derives prognostic information exclusively from the diagnostic angiogram without requiring patient demographics, comorbidities, biomarkers, or procedural variables, enabling risk stratification at initial coronary assessment. These results underscore the value of foundation model pretraining: the contrastive learning objective produces general-purpose representations that encode prognostically relevant visual phenotypes beyond discrete stenosis measurements, enabling transfer to downstream tasks not explicitly optimized during training and potentially complementing existing clinical risk scores.

For clinical deployment, we deployed DeepCORO-CLIP within a PACS-integrated inference pipeline (PACS-AI)[30,31] achieving <5 second processing times, enabling near real-time pre-procedural assessment during PCI planning. In practice, this allows operators to receive immediate stenosis quantification, lesion characterization, and preliminary risk stratification from the diagnostic angiogram before proceeding to intervention, supporting more informed revascularization decisions particularly in high-volume or single-operator settings where interpretive variability is highest. The architecture's hierarchical attention design, combining intra-video temporal modeling with cross-video multi-view aggregation, provides a unified mechanism for integrating information both within and across angiographic projections, mirroring the clinical workflow where operators mentally synthesize multiple views to form a composite assessment. This positions DeepCORO-CLIP as a general-purpose angiographic backbone that can be fine-tuned for additional clinical tasks beyond those evaluated here, including automated SYNTAX score estimation to guide revascularization strategy, preliminary angiographic report generation to reduce documentation burden, and longitudinal disease surveillance through serial embedding comparison.

Several limitations warrant consideration. First, the retrospective design may not fully capture the complexity of real-world clinical decision-making, and prospective validation is needed to assess the impact on procedural planning and temporal stability of model performance. Second, patients with prior CABG were excluded, limiting applicability to these populations. Third, while external validation demonstrated generalizability across one center, broader validation across diverse ethnic populations and healthcare systems with varying acquisition protocols remains necessary. Fourth, the study does not compare against clinically relevant alternatives including fractional flow reserve (FFR) or intra-vascular imaging. Fifth, video-text alignment relies on report-derived labels with inherent noise and potential misalignment between video-level content and narrative descriptions; robust handling of label noise was not formally addressed. Sixth, model performance depends on adequate video quality, which may limit applicability in resource-constrained settings or for suboptimal acquisitions common in clinical practice, yet our work achieves the highest performance compared to previous literature. Regarding MACE prediction specifically, although discriminatory performance exceeded clinically used models, two additional limitations should be noted. Negative controls were defined as patients with documented follow-up within 7 to 365 days without events; patients lacking follow-up were excluded rather than treated as censored observations, potentially introducing selection bias and limiting generalizability. Additionally, the model relies solely on angiographic video data as input. Incorporating clinical variables such as baseline comorbidities, laboratory values, medication history, and procedural characteristics into a multimodal framework could improve predictive performance and represents a direction for future work. We also did not do a head-to-head comparison with pre-existing models for this purpose. Future directions include interpretability mechanisms to enhance clinical acceptance, automated quality assessment to support iterative model improvement, intra-procedural guidance applications, and integration with invasive imaging modalities such as intravascular ultrasound and optical coherence tomography.

DeepCORO-CLIP demonstrates that video-text contrastive pretraining on coronary angiography produces general-purpose representations that transfer effectively across diagnostic, prognostic, and longitudinal monitoring tasks. By integrating multiple angiographic views into a unified embedding space, the model captures coronary anatomy and pathology at a level exceeding single-reader interpretation. With publicly released code, weights, and deployment infrastructure, DeepCORO-CLIP provides a foundation for automated angiographic interpretation and a building block for multimodal cardiovascular AI systems.

**Figure 1. DeepCORO-CLIP: A vision-language foundation model for coronary angiography interpretation.**

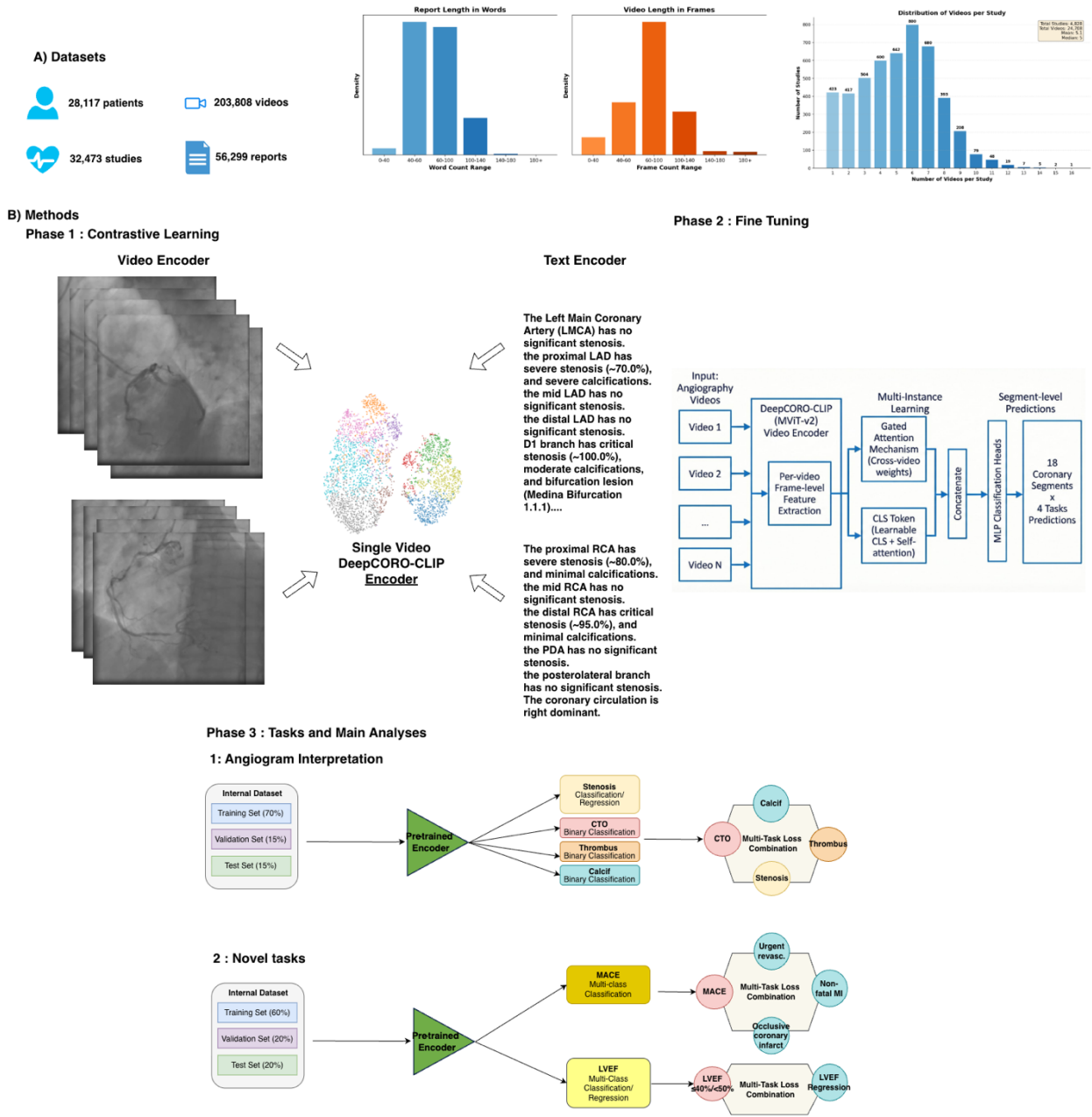

**Legend. (A) Dataset characteristics.** The internal cohort comprised 28,117 patients with 32,473 studies, yielding 203,808 angiography videos paired with 56,299 radiology reports. Histograms illustrate the distribution of report length, video frame counts, and number of videos per study. **(B) Model architecture and training phases.** Phase 1 employs contrastive learning to align video embeddings from the MViT-v2 encoder with corresponding text descriptions from catheterization reports. Phase 2 fine-tunes the pretrained encoder using a multi-instance learning framework with gated attention to aggregate frame-level features across multiple videos, generating segment-level predictions for 18 coronary segments across 4 tasks. **(C) Downstream tasks and analyses.** Task 1 evaluates angiogram interpretation including stenosis severity, CTO detection, thrombus identification, and calcification grading using multi-task learning (70/15/15 train/validation/test split). Task 2 assesses novel predictive capabilities for MACE (urgent revascularization, non-fatal MI, occlusive coronary infarct) and LVEF estimation using an independent split (60/20/20).

**Abbreviations:** MViT-v2, Multiscale Vision Transformer version 2; CLS, classification token; MLP, multi-layer perceptron; CTO, chronic total occlusion; Calcif, calcification; MACE, major adverse cardiovascular events; MI, myocardial infarction; LVEF, left ventricular ejection fraction; revasc., revascularization; LMCA, left main coronary artery; LAD, left anterior descending artery; RCA, right coronary artery; PDA, posterior descending artery.

**Figure 2. Results**

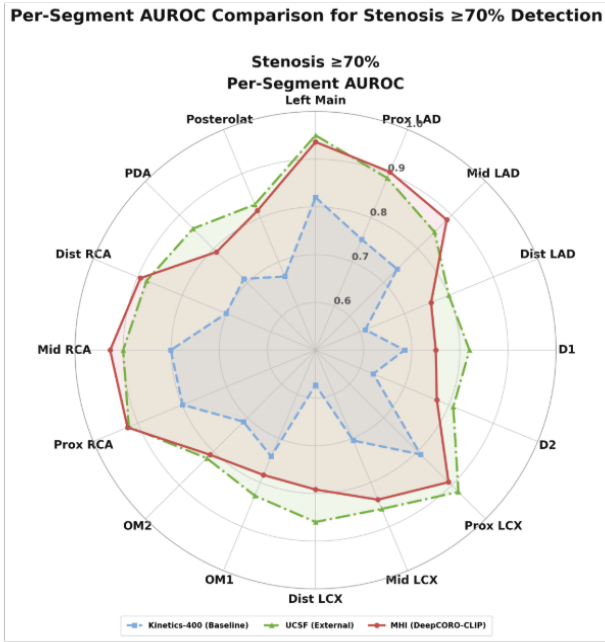
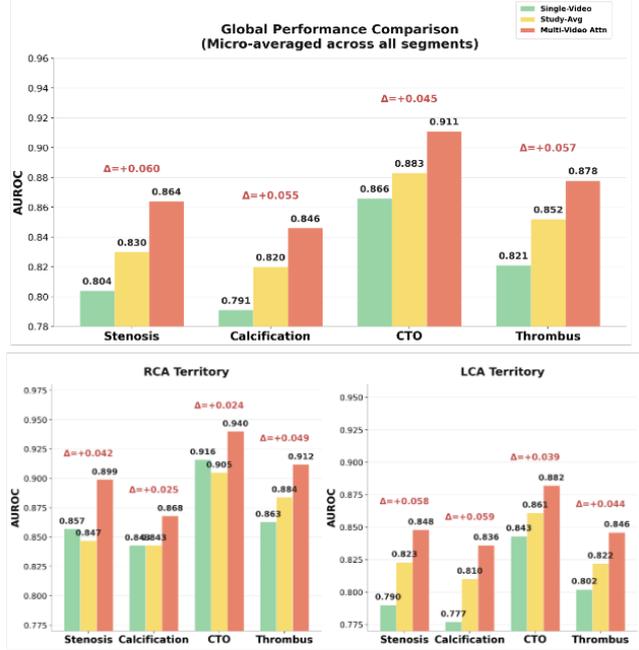
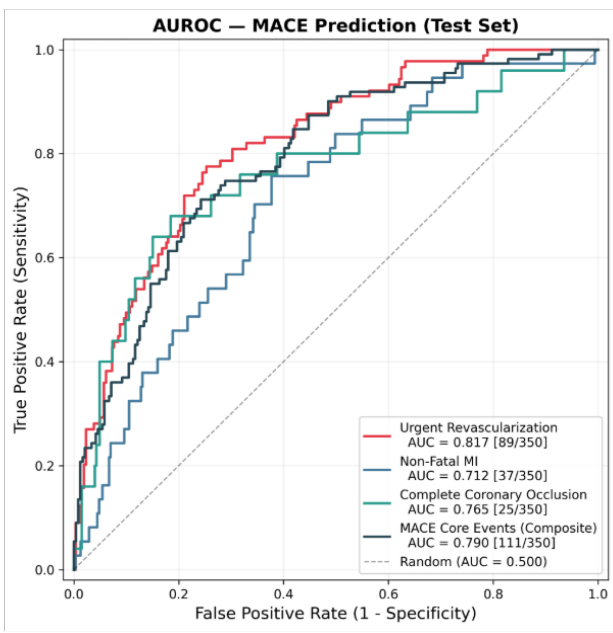
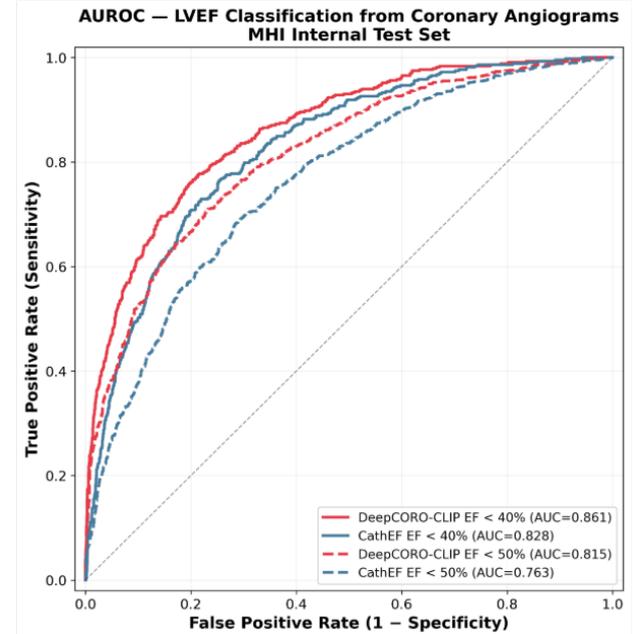

**Figure 2. DeepCORO-CLIP performance for stenosis detection, ablation studies, and transfer learning.**

**(A) Per-segment stenosis detection.** AUROC for ≥70% stenosis across 18 coronary segments comparing Kinetics-400 baseline, UCSF external validation, and MHI internal validation with DeepCORO-CLIP pretraining. **(B) Multi-view ablation experiments.** Global performance (micro-averaged) for stenosis, calcification, CTO, and thrombus detection comparing single-video, study-level averaging, and multi-video attention strategies. Delta values indicate improvement over single-video baseline. Bottom panels show territory-specific performance for RCA and LCA. **(C)**

**Transfer learning.** ROC curves for MACE prediction (left) and LVEF estimation compared against CathEF (right). **Abbreviations:** AUROC, area under the receiver operating characteristic curve; AUC, area under the curve; LAD, left anterior descending artery; LCX, left circumflex artery; RCA, right coronary artery; PDA, posterior descending artery; OM, obtuse marginal; CTO, chronic total occlusion; LCA, left coronary artery; MACE, major adverse cardiovascular events; MI, myocardial infarction; LVEF, left ventricular ejection fraction; EF, ejection fraction; MHI, Montreal Heart Institute; UCSF, University of California San Francisco.

**Table 1: Baseline Patient and Study Characteristics**

| Characteristic | Training Dataset | Validation Dataset | Test Dataset | p-value |
|---|---|---|---|---|
| Number of studies | 22,764 | 4,881 | 4,828 | N/A |
| Number of patients | 19,681 | 4,217 | 4,219 | N/A |
| Number of videos | 142,766 | 30,774 | 30,268 | N/A |
| **Demographics** | | | | |
| Mean age (years) ± STD | 68.2 ± 11.4 | 68.0 ± 11.1 | 68.7 ± 11.5 | 0.43 |
| **Sex** | | | | |
| Male, n (%) | 77,224 (65.9%) | 16,474 (65.1%) | 16,292 (65.9%) | 0.06 |
| Female, n (%) | 37,639 (32.1%) | 8,275 (32.7%) | 7,842 (31.7%) | 0.06 |
| **Medical History** | | | | |
| Hypertension, n (%) | 67,162 (57.3%) | 14,328 (56.6%) | 13,881 (56.2%) | 0.001 |
| Diabetes, n (%) | 26,265 (22.4%) | 5,431 (21.5%) | 5,468 (22.1%) | 0.003 |
| Chronic Kidney Disease, n (%) | 7,149 (6.1%) | 1,550 (6.1%) | 1,366 (5.5%) | 0.002 |
| Atrial Fibrillation, n (%) | 11,226 (9.6%) | 2,494 (9.9%) | 2,436 (9.9%) | 0.23 |
| Current or previous smoking, n (%) | 20,497 (17.5%) | 4,066 (16.1%) | 4,355 (17.6%) | <0.001 |
| Previous PCI, n (%) | 34 378 (29.4%) | 7 234 (28.6%) | 7 148 (28.9%) | 0.03 |
| **Video characteristics** | | | | |
| Frames per second, mean ± STD | 13.8 ± 2.9 | 13.8 ± 2.9 | 13.6 ± 3.0 | <0.001 |
| Number of frames per study, mean ± STD | 79.4 ± 29.6 | 79.8 ± 30.5 | 78.1 ± 26.9 | 0.68 |
| Views per study, mean ± STD | 5.1 ± 2.4 | 5.2 ± 2.5 | 5.1 ± 2.5 | 0.47 |

**Abbreviations:** STD, Standard Deviation; PCI, Percutaneous Coronary Intervention; MI, Myocardial Infarction; N/A, Not Applicable.

**Table 2. Consolidated Performance Metrics by Coronary Segment With 95% Confidence Intervals**

| Task | Metric | Global (All) | LCA | RCA |
|---|---|---|---|---|
| **Stenosis (Regression)** | n (abnormal/total) | 20,844 / 86,544 (24.1%) | 13,834 / 62,504 (22.1%) | 7,010 / 24,040 (29.2%) |
| | MAE (95% CI) | 9.241 (9.115–9.372) | 8.988 (8.837–9.135) | 9.899 (9.648–10.139) |
| | Pearson r (95% CI) | 0.525 (0.516–0.533) | 0.488 (0.477–0.499) | 0.590 (0.575–0.606) |
| **Stenosis ≥70%** | n (abnormal/total) | 7,270 / 86,544 (8.4%) | 5,024 / 62,504 (8.0%) | 2,246 / 24,040 (9.3%) |
| | AUROC (95% CI) | 0.888 (0.884–0.891) | 0.880 (0.876–0.885) | 0.905 (0.899–0.911) |
| | AUPRC (95% CI) | 0.517 (0.504–0.528) | 0.483 (0.469–0.499) | 0.590 (0.567–0.614) |
| | Sens (95% CI) | 0.847 (0.838–0.855) | 0.829 (0.818–0.840) | 0.887 (0.874–0.900) |
| | Spec (95% CI) | 0.702 (0.699–0.705) | 0.700 (0.696–0.704) | 0.706 (0.700–0.712) |
| **Calcification** | n (abnormal/total) | 1,685 / 86,544 (1.9%) | 1,141 / 62,504 (1.8%) | 544 / 24,040 (2.3%) |
| | AUROC (95% CI) | 0.903 (0.897–0.909) | 0.900 (0.893–0.908) | 0.907 (0.898–0.916) |

| Task | Metric | Global (All) | LCA | RCA |
|---|---|---|---|---|
| CTO | AUPRC (95% CI) | 0.181 (0.166–0.199) | 0.184 (0.165–0.205) | 0.184 (0.160–0.214) |
| | Sens (95% CI) | 0.847 (0.831–0.865) | 0.829 (0.807–0.849) | 0.886 (0.860–0.910) |
| | Spec (95% CI) | 0.804 (0.802–0.807) | 0.810 (0.807–0.814) | 0.789 (0.783–0.794) |
| | n (abnormal/total) | 550 / 86,544 (0.6%) | 273 / 62,504 (0.4%) | 277 / 24,040 (1.2%) |
| | AUROC (95% CI) | 0.923 (0.912–0.934) | 0.898 (0.879–0.915) | 0.946 (0.933–0.958) |
| | AUPRC (95% CI) | 0.179 (0.152–0.213) | 0.095 (0.072–0.127) | 0.265 (0.222–0.318) |
| | Sens (95% CI) | 0.836 (0.808–0.865) | 0.762 (0.714–0.811) | 0.910 (0.876–0.942) |
| | Spec (95% CI) | 0.860 (0.858–0.862) | 0.869 (0.867–0.872) | 0.835 (0.831–0.840) |
| Thrombus | n (abnormal/total) | 217 / 86,544 (0.3%) | 113 / 62,504 (0.2%) | 104 / 24,040 (0.4%) |
| | AUROC (95% CI) | 0.898 (0.874–0.923) | 0.880 (0.845–0.914) | 0.923 (0.888–0.954) |
| | AUPRC (95% CI) | 0.226 (0.168–0.294) | 0.171 (0.103–0.255) | 0.300 (0.211–0.404) |
| | Sens (95% CI) | 0.737 (0.681–0.796) | 0.611 (0.520–0.697) | 0.875 (0.806–0.932) |
| | Spec (95% CI) | 0.918 (0.916–0.920) | 0.952 (0.950–0.953) | 0.832 (0.827–0.836) |

**Abbreviations:** LCA, left coronary artery; RCA, right coronary artery; MAE, mean absolute error; AUROC, area under the receiver operating characteristic curve; AUPRC, area under the precision-recall curve; Sens, sensitivity; Spec, specificity; CTO, chronic total occlusion. Global (All) metrics represent sample-weighted averages across LCA and RCA territories.

Table 3. RCA and LCA Lesion Detection Metrics by Age and Gender Subgroups

| | Prevalence Patient (%) | Prevalence Lesion (%) | AUROC* (95% CI) | AUPRC* (95% CI) | Sensitivity* (95% CI) | Specificity* (95% CI) |
|---|---|---|---|---|---|---|
| | STENOSIS ≥ 70% | | | | | |
| Male | 3149 / 4828 (65.2 %) | 5637 / 56682 (9.94 %) | 0.87 (0.87 - 0.88) | 0.52 (0.50 - 0.53) | 0.82 (0.81 - 0.83) | 0.74 (0.74 - 0.75) |
| Female | 1559 / 4828 (32.3 %) | 1607 / 28062 (5.73 %) | 0.91 (0.90 - 0.92) | 0.52 (0.49 - 0.54) | 0.85 (0.83 - 0.87) | 0.81 (0.81 - 0.82) |
| Age ≤50 | 401 / 4828 (8.3 %) | 347 / 7218 (4.81 %) | 0.92 (0.91 - 0.93) | 0.48 (0.42 - 0.53) | 0.87 (0.84 - 0.91) | 0.83 (0.82 - 0.84) |
| Age 51-60 | 971 / 4828 (20.1 %) | 1228 / 17478 (7.03 %) | 0.90 (0.89 - 0.90) | 0.53 (0.50 - 0.56) | 0.83 (0.81 - 0.85) | 0.79 (0.78 - 0.80) |
| Age 61-70 | 1471 / 4828 (30.5 %) | 2257 / 26478 (8.52 %) | 0.89 (0.88 - 0.89) | 0.51 (0.49 - 0.53) | 0.87 (0.86 - 0.89) | 0.73 (0.73 - 0.74) |
| Age 71-80 | 1435 / 4828 (29.7 %) | 2424 / 25830 (9.38 %) | 0.88 (0.87 - 0.88) | 0.52 (0.50 - 0.54) | 0.80 (0.79 - 0.82) | 0.78 (0.77 - 0.78) |
| Age ≥80 | 550 / 4828 (11.4 %) | 1056 / 9900 (10.67 %) | 0.86 (0.85 - 0.87) | 0.52 (0.49 - 0.55) | 0.74 (0.72 - 0.77) | 0.80 (0.80 - 0.81) |
| | CALCIFICATION | | | | | |
| Male | 3149 / 4828 (65.2 %) | 1251 / 56682 (2.21 %) | 0.89 (0.88 - 0.90) | 0.18 (0.17 - 0.20) | 0.84 (0.82 - 0.86) | 0.80 (0.80 - 0.80) |
| Female | 1559 / 4828 (32.3 %) | 460 / 28062 (1.64 %) | 0.92 (0.91 - 0.93) | 0.19 (0.16 - 0.23) | 0.93 (0.91 - 0.96) | 0.76 (0.75 - 0.76) |
| Age ≤50 | 401 / 4828 (8.3 %) | 27 / 7218 (0.37 %) | 0.92 (0.87 - 0.96) | 0.15 (0.05 - 0.30) | 0.82 (0.67 - 0.96) | 0.90 (0.89 - 0.91) |
| Age 51-60 | 971 / 4828 (20.1 %) | 155 / 17478 (0.89 %) | 0.91 (0.90 - 0.93) | 0.14 (0.10 - 0.19) | 0.88 (0.82 - 0.93) | 0.78 (0.78 - 0.79) |
| Age 61-70 | 1471 / 4828 (30.5 %) | 445 / 26478 (1.68 %) | 0.89 (0.88 - 0.91) | 0.15 (0.12 - 0.18) | 0.86 (0.82 - 0.89) | 0.78 (0.77 - 0.78) |
| Age 71-80 | 1435 / 4828 (29.7 %) | 743 / 25830 (2.88 %) | 0.89 (0.88 - 0.90) | 0.22 (0.19 - 0.24) | 0.85 (0.82 - 0.87) | 0.80 (0.79 - 0.80) |
| Age ≥80 | 550 / 4828 (11.4 %) | 354 / 9900 (3.58 %) | 0.87 (0.85 - 0.89) | 0.22 (0.19 - 0.26) | 0.83 (0.79 - 0.86) | 0.76 (0.76 - 0.77) |
| | **THROMBUS** | | | | | |
| Male | 3149 / 4828 | 156 / 56682 | 0.89 (0.86 - 0.92) | 0.19 (0.13 - 0.27) | 0.72 (0.65 - 0.79) | 0.92 (0.91 - 0.92) |
|  | (65.2 %) | (0.28 %) | | | | |
| Female | 1559 / 4828 (32.3 %) | 55 / 28062 (0.20 %) | 0.91 (0.85 - 0.96) | 0.33 (0.19 - 0.46) | 0.73 (0.60 - 0.84) | 0.97 (0.97 - 0.97) |
| Age ≤50 | 401 / 4828 (8.3 %) | 27 / 7218 (0.37 %) | 0.91 (0.84 - 0.97) | 0.25 (0.10 - 0.45) | 0.81 (0.65 - 0.96) | 0.92 (0.91 - 0.93) |
| Age 51-60 | 971 / 4828 (20.1 %) | 62 / 17478 (0.35 %) | 0.92 (0.88 - 0.96) | 0.32 (0.21 - 0.45) | 0.79 (0.69 - 0.89) | 0.94 (0.93 - 0.94) |
| Age 61-70 | 1471 / 4828 (30.5 %) | 69 / 26478 (0.26 %) | 0.88 (0.83 - 0.93) | 0.24 (0.14 - 0.34) | 0.73 (0.62 - 0.83) | 0.92 (0.91 - 0.92) |
| Age 71-80 | 1435 / 4828 (29.7 %) | 51 / 25830 (0.20 %) | 0.89 (0.83 - 0.94) | 0.18 (0.08 - 0.30) | 0.67 (0.53 - 0.80) | 0.96 (0.96 - 0.96) |
| Age ≥80 | 550 / 4828 (11.4 %) | 10 / 9900 (0.10 %) | 0.89 (0.79 - 0.98) | 0.25 (0.02 - 0.59) | 0.69 (0.38 - 1.00) | 0.95 (0.95 - 0.96) |
| | **CTO** | | | | | |
| Male | 3149 / 4828 (65.2 %) | 430 / 56682 (0.76 %) | 0.91 (0.90 - 0.92) | 0.18 (0.15 - 0.22) | 0.85 (0.82 - 0.88) | 0.83 (0.83 - 0.83) |
| Female | 1559 / 4828 (32.3 %) | 119 / 28062 (0.42 %) | 0.95 (0.93 - 0.96) | 0.19 (0.13 - 0.25) | 0.95 (0.91 - 0.98) | 0.79 (0.79 - 0.80) |
| Age ≤50 | 401 / 4828 (8.3 %) | 21 / 7218 (0.29 %) | 0.96 (0.94 - 0.98) | 0.18 (0.06 - 0.34) | 0.91 (0.76 - 1.00) | 0.91 (0.91 - 0.92) |
| Age 51-60 | 971 / 4828 (20.1 %) | 103 / 17478 (0.59 %) | 0.93 (0.91 - 0.95) | 0.23 (0.15 - 0.31) | 0.91 (0.85 - 0.96) | 0.81 (0.80 - 0.82) |
| Age 61-70 | 1471 / 4828 (30.5 %) | 176 / 26478 (0.66 %) | 0.92 (0.90 - 0.94) | 0.18 (0.13 - 0.24) | 0.82 (0.76 - 0.88) | 0.89 (0.88 - 0.89) |
| Age 71-80 | 1435 / 4828 (29.7 %) | 164 / 25830 (0.63 %) | 0.93 (0.90 - 0.95) | 0.20 (0.14 - 0.26) | 0.83 (0.77 - 0.88) | 0.89 (0.88 - 0.89) |
| Age ≥80 | 550 / 4828 (11.4 %) | 92 / 9900 (0.93 %) | 0.88 (0.83 - 0.92) | 0.17 (0.11 - 0.25) | 0.82 (0.74 - 0.90) | 0.82 (0.81 - 0.83) |

**Abbreviations:** *: micro analysis; RCA: Right Coronary Artery; LCA: Left Coronary Artery; Prevalence Patient: Percentage of patients in each demographic group; Prevalence Lesion: Number of lesion divided by the total number of arterial segments within each demographic group; CTO: Chronic Total Occlusion; AUROC: Area Under the Receiver Operating Characteristic Curve; AUPRC: Area Under the Precision-Recall Curve; CI: Confidence Interval;

**Table 4. RCA and LCA Stenosis Regression Metrics by Age and Gender Subgroups**

| Subgroup | Prevalence (%) | MAE (95% CI) | Pearson (95% CI) |
|---|---|---|---|
| Male | 3149 / 4828 (65.2 %) | 10.68 (10.52 - 10.85) | 0.51 (0.50 - 0.52) |
| Female | 1559 / 4828 (32.3 %) | 6.98 (6.77 - 7.17) | 0.55 (0.53 - 0.57) |
| Age ≤50 | 401 / 4828 (8.3 %) | 5.78 (5.42 - 6.16) | 0.51 (0.47 - 0.55) |
| Age 51-60 | 971 / 4828 (20.1 %) | 8.12 (7.84 - 8.40) | 0.53 (0.51 - 0.55) |
| Age 61-70 | 1471 / 4828 (30.5 %) | 9.54 (9.29 - 9.77) | 0.51 (0.50 - 0.53) |
| Age 71-80 | 1435 / 4828 (29.7 %) | 10.22 (9.96 - 10.46) | 0.52 (0.51 - 0.54) |
| Age > 80 | 550 / 4828 (11.4 %) | 11.35 (10.92 - 11.78) | 0.53 (0.50 - 0.55) |

**Abbreviations:** RCA: Right Coronary Artery; LCA: Left Coronary Artery; Prevalence: Number of patients in each demographic group; MAE: Mean Absolute Error; CI: Confidence Interval; Pearson: Pearson correlation coefficient

**Table 5. Classification Performance for MACE Prediction (Test Set, n = 350)**

| MACE Label | n (abnormal/total) | AUROC (95% CI) | AUPRC (95% CI) | Sensitivity (95% CI) | Specificity (95% CI) |
|---|---|---|---|---|---|
| Urgent Revascularization | 89/350 (25.4%) | 0.818 (0.766-0.867) | 0.599 (0.498-0.707) | 0.78 (0.67-0.87) | 0.75 (0.66-0.84) |

| MACE Label | n (abnormal/total) | AUROC (95% CI) | AUPRC (95% CI) | Sensitivity (95% CI) | Specificity (95% CI) |
|---|---|---|---|---|---|
| Non-fatal MI | 37/350 (10.5%) | 0.712 (0.626-0.795) | 0.216 (0.144-0.344) | 0.76 (0.56-0.95) | 0.62 (0.43-0.86) |
| Complete Coronary Occlusion | 25/350 (7.1%) | 0.765 (0.640-0.870) | 0.273 (0.145-0.443) | 0.68 (0.48-0.89) | 0.82 (0.63-0.92) |
| Composite MACE | 111/350 (31.7%) | 0.790 (0.742-0.841) | 0.648 (0.557-0.734) | 0.71 (0.62-0.92) | 0.76 (0.53-0.84) |

**Abbreviations:** MACE: Major Adverse Cardiovascular Events; CI: Confidence Interval; HR: Hazard Ratio; Int: Intermediate; n: number (events/total); MI: Myocardial Infarction; n: number (abnormal/total)

# Extended Material

**Extended Figure 1. Flow Chart**

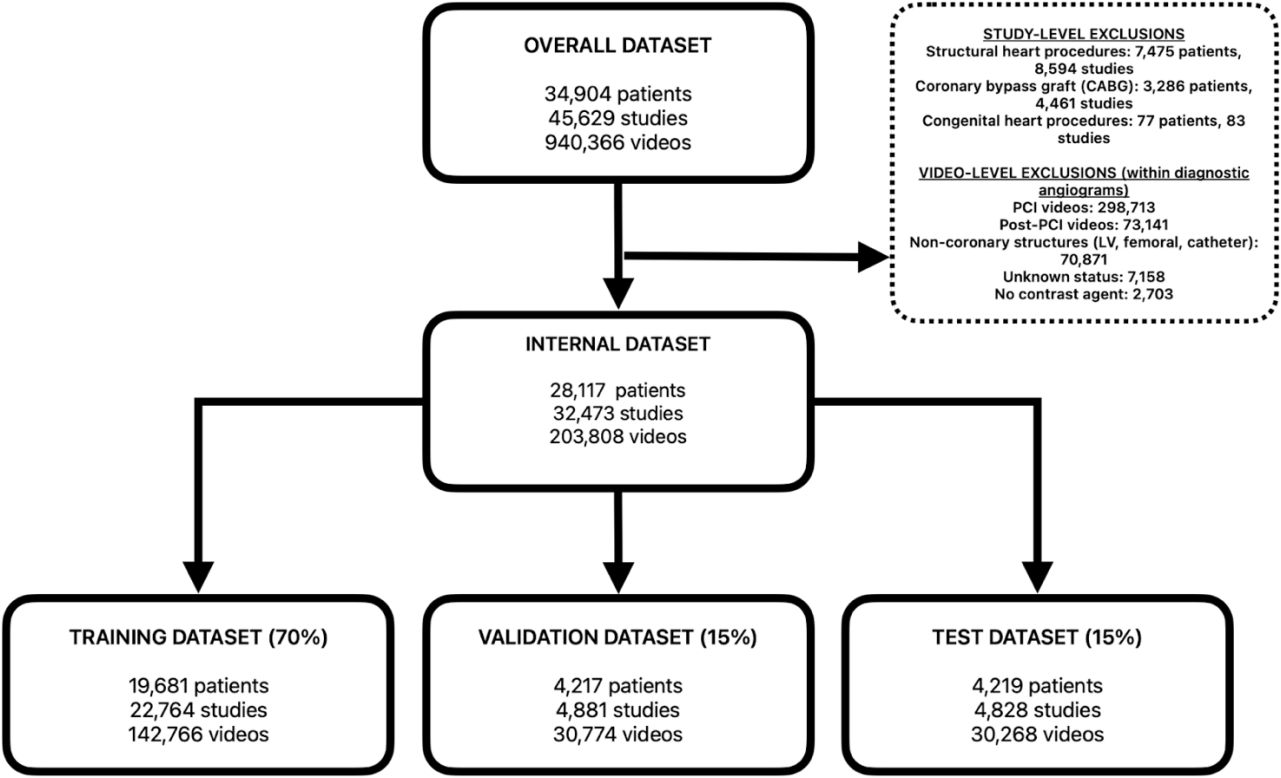

**Extended Figure 2. Sweep for best hyperparameters for the Video Encoder during Contrastive Learning Phase**

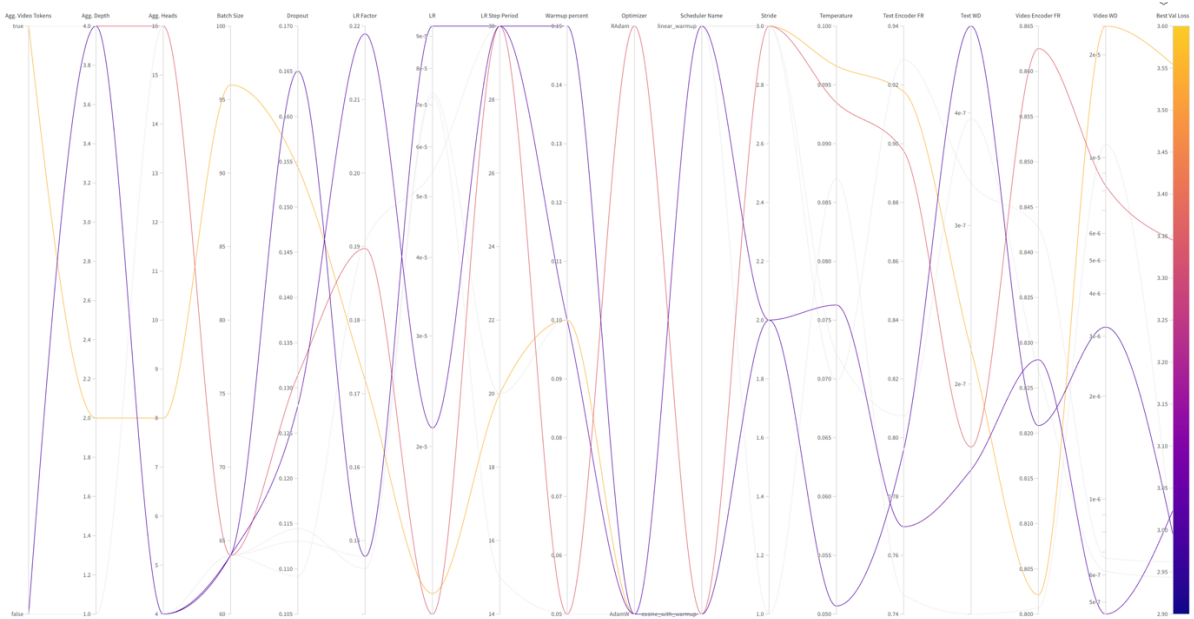

**Legend:** Systematic hyperparameter optimization across experimental runs identified temporal stride, text encoder freeze ratio, and contrastive temperature as the most influential parameters for model performance. Feature importance scores (left, blue) quantify the relative impact of each hyperparameter on validation loss, with stride showing the highest importance followed by text encoder freezing strategy. Correlation analysis (right) reveals the directional relationship between each parameter and validation loss, where positive correlations (teal) indicate parameter increases that degrade performance and negative correlations (pink) indicate parameter increases that improve performance. Learning rate exhibited strong negative correlation with validation loss, indicating that lower learning rates within the explored range ($5.8\times10^{-6}$ to $8.0\times10^{-5}$) improved contrastive alignment. Conversely, temporal stride, text freeze ratio, and temperature showed positive correlations, suggesting optimal performance at lower values within their respective ranges. Minor importance parameters including dropout, batch size, and video token aggregation strategy showed minimal influence on final model performance.

**Abbreviations:** lr, learning rate; num_heads: number of attention heads in multi-instance learning module.

## Extended Figure 3. Performance against other training objectives and architectures

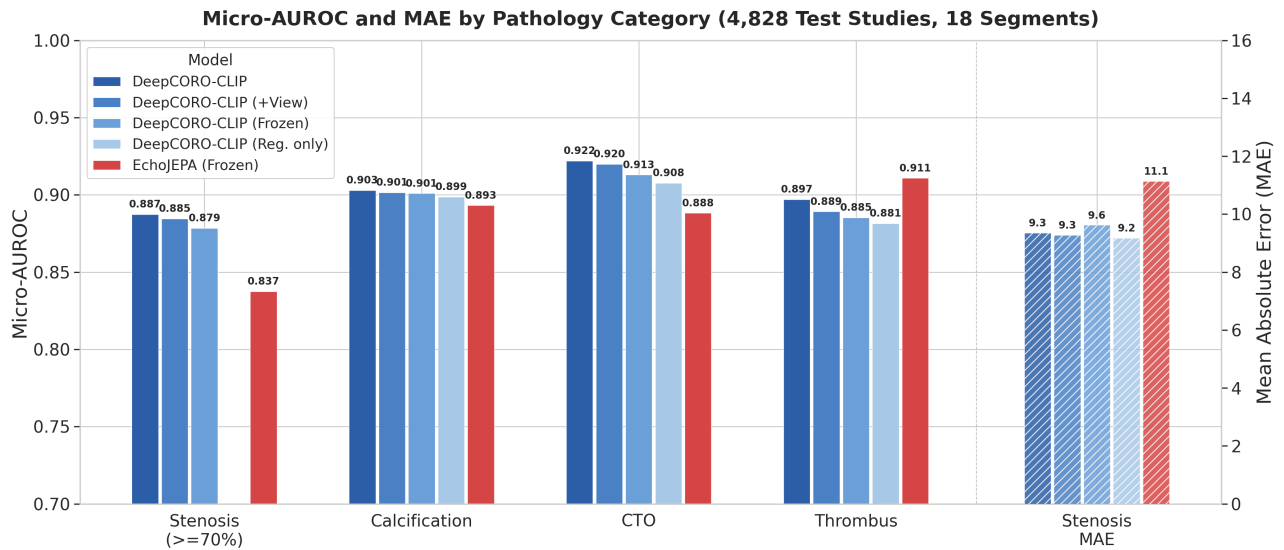

**Abbreviations:** MAE, Mean Absolute Error; AUROC, Area Under the Receiver Operating Characteristic Curve; CTO, Chronic Total Occlusion; Reg., Regression only.

## Extended Figure 4. DeepCORO-Clip Video Encoder Embedding analysis

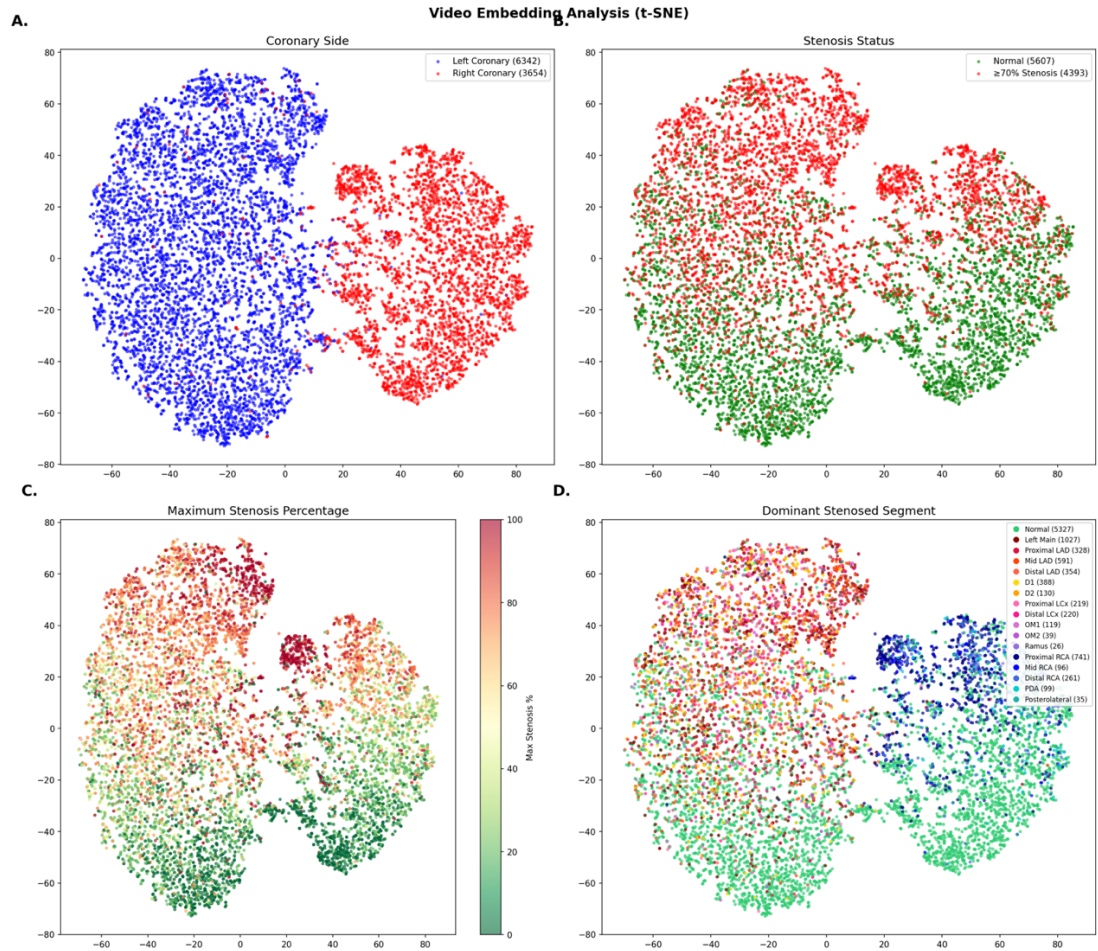

**FIGURE LEGEND:** t-SNE visualization of video embeddings colored by clinical attributes. (A) Coronary artery side classification showing separation between left and right coronary angiograms. (B) Binary stenosis status (normal vs ≥70% stenosis). (C) Maximum stenosis percentage displayed as a continuous color gradient. (D) Dominant stenosed coronary segment showing anatomical distribution of disease.

**Abbreviations**: LAD, left anterior descending artery; LCx, left circumflex artery; RCA, right coronary artery; OM, obtuse marginal branch; PDA, posterior descending artery; D1/D2, first and second diagonal branches; t-SNE, t-distributed stochastic neighbor embedding.

**Extended Figure 5. Study-level embedding distance reflects longitudinal changes in coronary artery disease status**

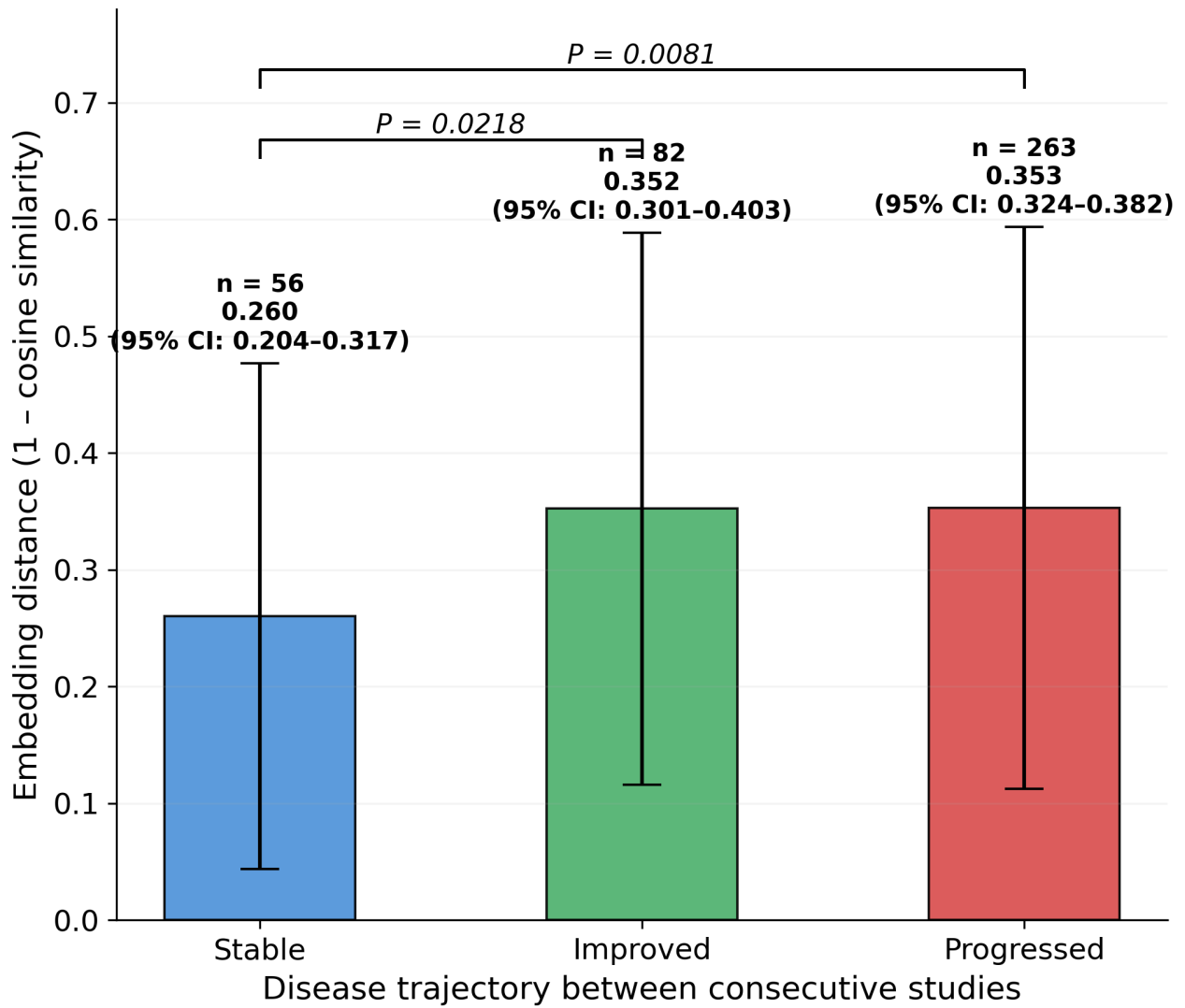

**Legend:** Embedding distance (1 − cosine similarity) between consecutive coronary angiography studies stratified by disease trajectory. Disease status was classified as stable (no vessel with ≥20% absolute change in stenosis), improved (≥1 vessel with ≥20% decrease in stenosis without concurrent progression), or progressed (≥1 vessel developing new significant stenosis >50% or worsening by ≥20%). When PCI was performed, the post-procedural embedding was used to represent the patient's coronary anatomy at study transition. Error bars indicate standard deviation. P-values from two-sample t-tests. n = 401 consecutive study pairs from 352 patients.
**Abbreviations:** PCI, percutaneous coronary intervention; CI, confidence interval; SD, standard deviation.

**Extended Figure 6. PACS-AI Deployment of DeepCORO-CLIP**

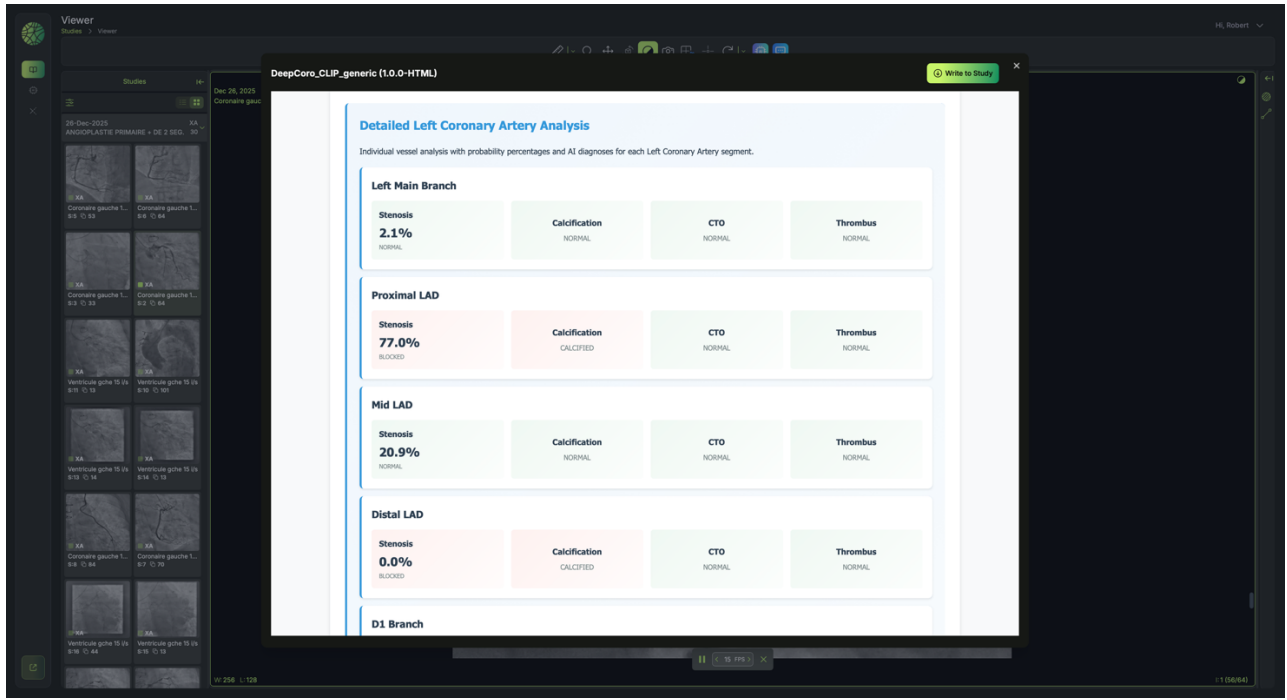

## Extended Table 1. Class Definitions of Angiographic Projection Angle Used for View Classification

| Class | Definition |
|---|---|
| RAO Cranial | -45° to -15° RAO; 15° to 45° Cranial |
| AP Cranial | -15° to 15° AP; 15° to 45° Cranial |
| LAO Cranial | 15° to 45° LAO; 15° to 45° Cranial |
| RAO Straight | -45° to -15° RAO; -15° to 15° AP |
| AP | -15° to 15° AP; -15° to 15° AP |
| RAO Caudal | -45° to -15° RAO; -45° to -15° Caudal |
| AP Caudal | -15° to 15° AP; -45° to -15° Caudal |
| LAO Caudal | 15° to 45° LAO; -45° to -15° Caudal |
| LAO Straight | 15° to 45° LAO; -15° to 15° AP |
| LAO Lateral | 70° to 110° LAO; -15° to 15° AP |
| RAO Lateral | -110° to -70° RAO; -15° to 15° AP |
| Other | Any angles not belonging to the previous definitions |

# Extended Table 2. Primary Structure Identification, Presence of Stenting Procedures and Contrast Presence

**Anatomic Structure Definitions**

| Structure | Definition |
|---|---|
| Left Coronary Artery | Artery that arises from the aorta above the left cusp of the aortic valve |
| Right Coronary Artery | Artery that arises from the aorta above the right cusp of the aortic valve |
| Other | Any images not belonging to defined classes (e.g., kidneys, pacemaker, peripheral vessels) |
| Bypass Graft | Venous graft, internal mammary artery graft, or radial artery graft |
| Catheter | Any guiding catheter or diagnostic catheter without other underlying structures And no contrast injection |
| Femoral Artery | Superficial, deep, or common femoral artery |
| Left Ventricle | Left ventricular cavity as delimited during ventriculography |
| TAVR | Transcatheter aortic valve replacement procedure or prosthesis |
| Aorta | Ascending aorta, aortic arch, or descending aorta as delimited during aortography |
| Radial Artery | Major artery in the forearm used for vascular access |
| TEE Probe | Transesophageal echocardiography probe |
| Pigtail Catheter | Pigtail catheter without other underlying structures |

**Additional Annotations**

| Annotation | Definition |
|---|---|
| Contrast Agent | Presence (yes) or absence (no) of contrast opacification in coronary vessels |
| Stent Presence | Presence (present) or absence (absent) of previously deployed coronary stents |
| Coronary Dominance | Right dominant (posterior descending artery arises from RCA) or left dominant (posterior descending artery arises from left circumflex) |

**Abbreviation:** TEE : Transesophageal echocardiography; TAVR: Transcatheter aortic valve replacement.

### Extended Table 3. Contrast Opacity, Main structure, Stent Presence and Coronary Dominance Performance on the Test Set (n=140 videos)

| Task | N | Prevalence (%) | AUROC (95% CI) | AUPRC (95% CI) | Sensitivity (95% CI) | Specificity (95% CI) |
|---|---|---|---|---|---|---|
| Contrast Opacity | 93 | 66.4 % | 0.95 (0.92 - 0.98) | 0.97 (0.94 - 0.99) | 0.91 (0.85 - 0.97) | 0.94 (0.86 - 1.00) |
| Main Structure* | 140 | 100% | 0.98 (0.96 - 0.99) | 0.90 (0.84 - 0.95) | 0.86 (0.81 - 0.91) | 0.99 (0.98 - 0.99) |
| Stent Presence | 42 | 37.1 % | 0.97 (0.94 - 0.99) | 0.94 (0.86 - 0.99) | 0.92 (0.84 - 0.98) | 0.92 (0.86 - 0.98) |
| Coronary Dominance | 99 | 29.3 % | 0.84 (0.76 - 0.91) | 0.69 (0.55 - 0.83) | 0.80 (0.68 - 0.92) | 0.82 (0.75 - 0.90) |

**Abbreviations**: *: micro analysis; N: number of patients; AUROC, area under the receiver operating characteristic curve; AUPRC, area under the precision-recall curve; CI, confidence interval.

**Extended Table 4: Per-class Performance Metrics for Main Structure Classification on Test Set (n=140 videos)**

| Class | N | Prevalence (%) | AUROC (95% CI) | AUPRC (95% CI) | Sensitivity (95% CI) | Specificity (95% CI) |
|---|---|---|---|---|---|---|
| Left Coronary | 46 | 32.9 | 0.99 (0.98-1.00) | 0.99 (0.97-1.00) | 0.96 (0.89-1.00) | 0.99 (0.97-1.00) |
| Right Coronary | 32 | 22.9 | 0.99 (0.98-1.00) | 0.98 (0.94-1.00) | 0.97 (0.90-1.00) | 0.97 (0.94-1.00) |
| Other | 18 | 12.9 | 0.96 (0.92-0.98) | 0.65 (0.46-0.87) | 0.94 (0.81-1.00) | 0.94 (0.90-0.98) |
| Graft | 9 | 6.4 | 0.98 (0.95-1.00) | 0.93 (0.73-1.00) | 0.89 (0.67-1.00) | 0.99 (0.98-1.00) |
| Catheter | 13 | 9.3 | 0.81 (0.70-0.91) | 0.42 (0.18-0.65) | 0.23 (0.00-0.50) | 0.99 (0.98-1.00) |
| Femoral | 5 | 3.6 | 1.0 (1.00-1.00) | 1.00 (1.00-1.00) | 1.00 (1.00-1.00) | 0.99 (0.98-1.00) |
| LV | 4 | 2.9 | 1.00 (1.00-1.00) | 1.00 (1.00-1.00) | 1.00 (1.00-1.00) | 1.00 (1.00-1.00) |
| TAVR | 3 | 2.1 | 1.00 (0.99-1.00) | 0.92 (0.50-1.00) | 0.67 (0.00-1.00) | 0.99 (0.98-1.00) |
| Aorta | 4 | 2.9 | 1.00 (0.99-1.00) | 0.95 (0.72-1.00) | 0.75 (0.00-1.00) | 0.99 (0.98-1.00) |
| Radial | 3 | 2.1 | 1.00 (0.99-1.00) | 0.92 (0.50-1.00) | 1.00 (0.00-1.00) | 0.99 (0.98-1.00) |
| TEE probe | 0 | 0.0 | N/A | N/A | N/A | 1.00 (1.00-1.00) |
| Pigtail | 3 | 2.1 | 0.95 (0.88-0.99) | 0.28 (0.07-0.83) | 0.33 (0.00-1.00) | 0.99 (0.96-1.00 |

**Abbreviations**: N: number of patients; AUROC, area under the receiver operating characteristic curve; AUPRC, area under the precision-recall curve; CI, confidence interval; LV: left ventricle; TAVR: transcatheter aortic valve replacement; TEE: transesophageal echocardiography

## Extended Table 5. Coronary Artery Segment Definitions According to SYNergy between percutaneous coronary intervention with TAXus and cardiac surgery Score

**Left Coronary Artery**

| Segment | Definition |
|---|---|
| Left Main (5) | From the ostium of the left coronary artery through bifurcation into left anterior descending and left circumflex branches |
| Proximal LAD (6) | Proximal to and including first major septal branch |
| Mid LAD (7) | LAD immediately distal to origin of first septal branch and extending to the point where LAD forms an angle (right anterior oblique view). If this angle is not identifiable, this segment ends at one half the distance from the first septal to the apex of the heart |
| Distal LAD (8) | Terminal portion of LAD, beginning at the end of segment 7 and extending to or beyond the apex |
| First Diagonal (D1) (9) | The first diagonal originating from segment 6 or 7 |
| Second Diagonal (D2) (10) | Originating from segment 8 or the transition between segment 7 and 8 |
| Proximal LCX (11) | Main stem of circumflex from its origin at left main and including origin of first obtuse marginal branch17 |
| Mid-Distal LCX (13) | The stem of the circumflex distal to the origin of the most distal obtuse marginal branch, running along the posterior left atrioventricular groove |
| Distal LCX (15) | Most distal part of dominant left circumflex when present, giving origin to septal branches. When this artery is present, segment 4 is usually absent |

| Segment | Definition |
|---|---|
| Left Ventricular Posterior (LVP) (14) | Running to the posterolateral surface of the left ventricle. May be absent or a division of obtuse marginal branch |
| Ramus Intermedius (RI) (12) | The ramus intermedius is a variant coronary artery arising from trifurcation of the left main, present in 10-30% of patients. It originates between the LAD and LCx, supplying the lateral and inferior walls in a distribution similar to diagonal or obtuse marginal branches. |
| First Obtuse Marginal (OM1) (12a) | First side branch of circumflex running in general to the area of obtuse margin of the heart |
| Second Obtuse Marginal (OM2) (12b) | Second branch of circumflex running in the same direction as OM1 |

**Right Coronary Artery**

| Segment | Definition |
|---|---|
| Proximal RCA (1) | From the ostium to one half the distance to the acute margin of the heart. |
| Mid RCA (2) | From the end of first segment to acute margin of heart. |
| Distal RCA (3) | From the acute margin of the heart to the origin of the posterior descending artery. |
| Posterior Descending Artery (PDA) (4) | Running in the posterior interventricular groove. |
| Posterolateral (16) | Posterolateral branch originating from the distal coronary artery distal to the crux. |

**Abbreviations:** LM: left main coronary artery; LAD: left anterior descending coronary artery; LCx: left circumflex coronary artery; RCA: right coronary artery; D1: first diagonal branch of the LAD; D2: second diagonal branch of the LAD; OM1: first obtuse marginal branch of the LCx; OM2: second obtuse marginal branch of the LCx; RI: ramus intermedius coronary artery; LVP: left ventricular posterior branch; PDA: posterior descending artery; PL: posterolateral branch.

## Extended Table 6. Range of Values Explored for Contrastive Learning

| Parameter | Range Explored | Best (Single) | Best (Multi MVIT) | Best (Multi X3D) |
|---|---|---|---|---|
| Model | Mvit2s_b, x3d_m | Mvit2s_b | Mvit2s | x3d_m |
| Number Videos | 1 - 8 | 1 | 4 | 4 |
| Loss Function | CLIP, InfoNCE, SIGLIP | CLIP, InfoNCE, SIGLIP | CLIP, InfoNCE, SIGLIP | CLIP, InfoNCE, SIGLIP |
| Frames | 16 - 64 | 16 | 16 | 32 |
| Stride | 1 - 3 | 2 | 2 | 1 |
| Epochs | 20 - 30 | 30 | 30 | 25 |
| Learning Rate | 5.8e-6 - 8.0e-5 | 7.2e-5 | 1e-5 | 1e-5 |
| Dropout | 0.10 - 0.19 | 0.11 | 0.10 | 0.10 |
| Temperature | 0.06 - 0.11 | 0.087 | 0.061 | 0.063 |
| Video Freeze % | 72.8% - 94.8% | 80.1% | 80% | 80% |
| Text Freeze % | 60.7% - 90.2% | 74.6% | 80% | 80% |
| Text Weight Decay | 1e-7 - 1e-5 | 1.1e-7 | 7.97e-7 | 2.10e-7 |
| Video Weight Decay | 1e-7 - 1e-4 | 1.09e-5 | 6.48e-6 | 2.20e-5 |
| Batch Size | 12 - 64 | 64 | 12 | 12 |

## Extended Table 7. Range of Values Explored for Transfer Learning

| Parameter | Range Explored | Best Value |
|---|---|---|
| **Training** | | |
| Epochs | 5 – 50 | 25 |
| Videos | 6 – 10 | 10 |
| Batch Size | {6, 8, 10, 12, 16} | 12 |
| Random Augmentation | {True, False} | True |
| **Video Encoder** | | |
| Freeze Ratio | 75 – 85% | 84.8% |
| Learning Rate | $2\times10^{-6} - 2\times10^{-5}$ | $5.9\times10^{-6}$ |
| Weight Decay | $0-5\times10^{-7}$ | $4.4\times10^{-7}$ |
| **Attention Mechanism** | | |
| Attention Heads | {6, 8} | 6 |
| Hidden Dimension | 256, 512 | 256 |
| Pooling Mode | Mean, Attention, CLS, Attention + CLS | Attention + CLS token |
| Dropout | 0.15 – 0.45 | 0.24 |
| Within-Video Attention LR | $5\times10^{-4} - 5\times10^{-3}$ | $1.0\times10^{-3}$ |
| Across-Video Attention LR | $5\times10^{-5} - 5\times10^{-3}$ | $1.93\times10^{-4}$ |
| Aggregation LR | $5\times10^{-5} - 5\times10^{-3}$ | $8.1\times10^{-5}$ |
| Within-Video Weight Decay | $1\times10^{-6} - 1\times10^{-4}$ | $2.6\times10^{-6}$ |
| Across-Video Weight Decay | $1\times10^{-6} - 1\times10^{-4}$ | $1.1\times10^{-5}$ |
| Aggregation Weight Decay | $1\times10^{-6} - 1\times10^{-3}$ | $1.4\times10^{-6}$ |
| **Learning Rate Schedule** | | |
| Scheduler | Linear with decay, Linear, Cosine, Cosine w/ warm restarts | Cosine w/ warm restarts |

**Abbreviations** : CLS : Classification.

| Extended Table 8. Recall metrics of the Contrastive Learning Training Phase | Video → Text (V2T) | Text → Video (T2V) |
| --- | --- | --- |
| **VIDEO-LEVEL METRICS** | | |
| Recall@1 | 6.70% | 1.11% |
| Recall@5 | 12.62% | 4.57% |
| Recall@10 | 16.70% | 7.18% |
| Recall@25 | 24.25% | 13.80% |
| Recall@50 | 31.95% | 21.02% |
| Mean Rank | 418.34 | 997.55 |
| Median Rank | 156 | 292 |
| **STUDY-LEVEL METRICS (Aggregated to 4,828 studies)** | | |
| Recall@1 | 5.34% | 1.74% |
| Recall@5 | 10.59% | 5.39% |
| Recall@10 | 14.26% | 8.90% |
| Recall@25 | 21.23% | 15.90% |
| Recall@50 | 28.08% | 22.93% |
| Mean Rank | 603.43 | 784.71 |
| Median Rank | 227 | 297 |

## Extended Table 9. Stenosis Quantification by Coronary Segment

| Segment | n ≥70% (%) | MAE (95% CI) | r (95% CI) | AUROC (95% CI) | AUPRC (95% CI) | Sens (95% CI) | Spec (95% CI) | Youden threshold |
|---|---|---|---|---|---|---|---|---|
| Left Main | 371/4828 (7.7%) | 8.42 (7.95-8.92) | 0.61 (0.58-0.64) | 0.937 (0.927-0.947) | 0.640 (0.587-0.686) | 0.89 (0.85-0.92) | 0.84 (0.83-0.85) | 0.0546 |
| Proximal LAD | 810/4828 (16.8%) | 15.81 (15.34-16.33) | 0.64 (0.62-0.67) | 0.903 (0.893-0.913) | 0.676 (0.643-0.708) | 0.86 (0.84-0.88) | 0.78 (0.77-0.79) | 0.1145 |
| Mid LAD | 1040/4828 (21.5%) | 18.02 (17.51-18.49) | 0.65 (0.62-0.67) | 0.885 (0.875-0.895) | 0.702 (0.673-0.728) | 0.83 (0.81-0.85) | 0.77 (0.76-0.79) | 0.1754 |
| Distal LAD | 160/4828 (3.3%) | 4.32 (3.86-4.76) | 0.01 (-0.03-0.03) | 0.754 (0.717-0.788) | 0.107 (0.075-0.145) | 0.67 (0.60-0.74) | 0.73 (0.71-0.74) | 0.0373 |
| D1 | 537/4828 (11.1%) | 14.57 (13.86-15.27) | 0.23 (0.20-0.26) | 0.756 (0.738-0.773) | 0.269 (0.237-0.303) | 0.75 (0.72-0.79) | 0.63 (0.62-0.65) | 0.1308 |
| D2 | 294/4828 (6.1%) | 7.58 (7.02-8.15) | 0.12 (0.09-0.15) | 0.769 (0.746-0.794) | 0.149 (0.126-0.175) | 0.82 (0.77-0.86) | 0.61 (0.60-0.63) | 0.0568 |
| RI | 170/4828 (3.5%) | 4.65 (4.19-5.09) | 0.12 (0.08-0.15) | 0.769 (0.738-0.798) | 0.099 (0.076-0.128) | 0.81 (0.75-0.87) | 0.62 (0.61-0.64) | 0.0299 |
| Proximal LCX | 548/4828 (11.4%) | 14.43 (13.94-14.94) | 0.56 (0.53-0.59) | 0.893 (0.881-0.905) | 0.527 (0.480-0.571) | 0.89 (0.86-0.92) | 0.74 (0.73-0.75) | 0.0830 |

| Segment | n ≥70% (%) | MAE (95% CI) | r (95% CI) | AUROC (95% CI) | AUPRC (95% CI) | Sens (95% CI) | Spec (95% CI) | Youden threshold |
|---|---|---|---|---|---|---|---|---|
| Mid LCX | 361/4828 (7.5%) | 9.82 (9.20-10.46) | 0.06 (0.03-0.09) | 0.841 (0.822-0.858) | 0.273 (0.233-0.315) | 0.86 (0.82-0.89) | 0.70 (0.68-0.71) | 0.0597 |
| Distal LCX | 90/4828 (1.9%) | 2.50 (2.15-2.85) | 0.12 (0.08-0.15) | 0.804 (0.758-0.848) | 0.119 (0.071-0.179) | 0.90 (0.83-0.95) | 0.57 (0.56-0.58) | 0.0105 |
| LVP | 23/4828 (0.5%) | 0.56 (0.40-0.76) | 0.00 (-0.03-0.03) | 0.930 (0.898-0.959) | 0.064 (0.029-0.123) | 1.00 (1.00-1.00) | 0.74 (0.73-0.75) | 0.0723 |
| OM1 | 307/4828 (6.4%) | 8.39 (7.81-8.98) | 0.19 (0.16-0.22) | 0.790 (0.768-0.812) | 0.185 (0.154-0.219) | 0.75 (0.69-0.80) | 0.71 (0.70-0.73) | 0.0767 |
| OM2 | 340/4828 (7.0%) | 8.91 (8.30-9.57) | 0.11 (0.08-0.14) | 0.813 (0.794-0.832) | 0.229 (0.193-0.266) | 0.82 (0.78-0.86) | 0.67 (0.66-0.68) | 0.0014 |
| Proximal RCA | 789/4828 (16.3%) | 14.24 (13.76-14.71) | 0.73 (0.71-0.75) | 0.923 (0.914-0.933) | 0.728 (0.695-0.761) | 0.87 (0.85-0.90) | 0.82 (0.81-0.83) | 0.1079 |
| Mid RCA | 714/4828 (14.8%) | 14.06 (13.48-14.60) | 0.65 (0.62-0.68) | 0.926 (0.917-0.934) | 0.685 (0.650-0.721) | 0.89 (0.87-0.91) | 0.80 (0.79-0.82) | 0.1284 |
| Distal RCA | 348/4828 (7.2%) | 10.27 (9.69-10.87) | 0.49 (0.46-0.52) | 0.897 (0.882-0.911) | 0.462 (0.408-0.516) | 0.96 (0.94-0.98) | 0.67 (0.66-0.69) | 0.0327 |

| Segment | n ≥70% (%) | MAE (95% CI) | r (95% CI) | AUROC (95% CI) | AUPRC (95% CI) | Sens (95% CI) | Spec (95% CI) | Youden threshold |
|---|---|---|---|---|---|---|---|---|
| PDA | 239/4828 (5.0%) | 6.93 (6.36-7.46) | 0.21 (0.17-0.24) | 0.795 (0.770-0.820) | 0.180 (0.141-0.225) | 0.87 (0.83-0.91) | 0.57 (0.56-0.59) | 0.0299 |
| Posterolateral | 171/4828 (3.5%) | 4.72 (4.27-5.22) | 0.22 (0.18-0.25) | 0.816 (0.786-0.846) | 0.142 (0.107-0.180) | 0.82 (0.75-0.88) | 0.68 (0.67-0.69) | 0.0209 |

**Abbreviations**: MAE, mean absolute error (%); r, Pearson correlation; AUROC, area under the receiver operating characteristic curve; AUPRC, area under the precision-recall curve; Sens, sensitivity; Spec, specificity; CI, confidence interval; LAD, left anterior descending; LCX, left circumflex; RI: Ramus Intermedius; RCA, right coronary artery; D1/D2, diagonal branches; OM1/OM2, obtuse marginal branches; LVP, left ventricular posterior branch; PDA, posterior descending artery.

## Extended Table 10. Calcification Quantification by Coronary Segment

| Segment | n (%) | AUROC (95% CI) | AUPRC (95% CI) | Sens (95% CI) | Spec (95% CI) | Youden threshold |
|---|---|---|---|---|---|---|
| Left Main | 164/4828 (3.4%) | 0.884 (0.862-0.905) | 0.262 (0.200-0.330) | 0.86 (0.80-0.91) | 0.77 (0.75-0.78) | 0.0230 |
| Proximal LAD | 320/4828 (6.6%) | 0.805 (0.780-0.829) | 0.232 (0.194-0.272) | 0.72 (0.67-0.77) | 0.76 (0.75-0.77) | 0.0729 |
| Mid LAD | 271/4828 (5.6%) | 0.820 (0.797-0.843) | 0.219 (0.182-0.262) | 0.72 (0.67-0.77) | 0.78 (0.76-0.79) | 0.0894 |
| Distal LAD | 15/4828 (0.3%) | 0.847 (0.760-0.921) | 0.037 (0.009-0.092) | 0.87 (0.67-1.00) | 0.72 (0.71-0.73) | 0.0026 |

| Segment | n (%) | AUROC (95% CI) | AUPRC (95% CI) | Sens (95% CI) | Spec (95% CI) | Youden threshold |
|---|---|---|---|---|---|---|
| D1 | 66/4828 (1.4%) | 0.806 (0.753-0.857) | 0.064 (0.040-0.096) | 0.73 (0.62-0.84) | 0.76 (0.75-0.77) | 0.0206 |
| D2 | 38/4828 (0.8%) | 0.880 (0.845-0.915) | 0.055 (0.029-0.098) | 1.00 (1.00-1.00) | 0.63 (0.62-0.65) | 0.0065 |
| RI | 25/4828 (0.5%) | 0.832 (0.767-0.890) | 0.024 (0.012-0.039) | 0.92 (0.79-1.00) | 0.61 (0.59-0.62) | 0.0029 |
| Proximal LCX | 151/4828 (3.1%) | 0.861 (0.837-0.882) | 0.191 (0.141-0.248) | 0.84 (0.78-0.90) | 0.74 (0.72-0.75) | 0.0207 |
| Mid LCX | 40/4828 (0.8%) | 0.842 (0.787-0.889) | 0.045 (0.024-0.074) | 0.90 (0.79-0.98) | 0.69 (0.67-0.70) | 0.0062 |
| Distal LCX | 10/4828 (0.2%) | 0.900 (0.801-0.977) | 0.049 (0.009-0.140) | 0.81 (0.50-1.00) | 0.88 (0.87-0.89) | 0.0050 |
| LVP | 4/4828 (0.1%) | 0.918 (0.822-0.983) | 0.013 (0.002-0.034) | 1.00 (1.00-1.00) | 0.81 (0.80-0.83) | 0.0104 |
| OM1 | 28/4828 (0.6%) | 0.871 (0.830-0.914) | 0.090 (0.023-0.206) | 0.86 (0.71-0.97) | 0.78 (0.77-0.80) | 0.0091 |
| OM2 | 33/4828 (0.7%) | 0.858 (0.795-0.908) | 0.054 (0.025-0.100) | 0.88 (0.75-0.97) | 0.70 (0.69-0.72) | - |
| Proximal RCA | 279/4828 (5.8%) | 0.853 (0.834-0.872) | 0.253 (0.210-0.302) | 0.87 (0.83-0.91) | 0.70 (0.69-0.72) | 0.0376 |
| Mid RCA | 169/4828 (3.5%) | 0.872 (0.850-0.893) | 0.196 (0.151-0.248) | 0.92 (0.88-0.96) | 0.71 (0.70-0.72) | 0.0302 |
| Distal RCA | 71/4828 (1.5%) | 0.900 (0.874-0.924) | 0.105 (0.071-0.144) | 0.87 (0.79-0.94) | 0.81 (0.80-0.82) | 0.0171 |
| PDA | 23/4828 (0.5%) | 0.874 (0.822-0.921) | 0.077 (0.023-0.165) | 0.83 (0.66-0.96) | 0.80 (0.79-0.81) | 0.0066 |

| Segment | n (%) | AUROC (95% CI) | AUPRC (95% CI) | Sens (95% CI) | Spec (95% CI) | Youden threshold |
|---|---|---|---|---|---|---|
| Posterolateral | 17/4828 (0.4%) | 0.852 (0.785-0.915) | 0.023 (0.009-0.049) | 1.00 (1.00-1.00) | 0.62 (0.60-0.63) | 0.0012 |

**Abbreviations**; n, abnormal/total segments (moderate or severe calcification); AUROC, area under the receiver operating characteristic curve; AUPRC, area under the precision-recall curve; Sens, sensitivity; Spec, specificity; LAD, left anterior descending; LCX, left circumflex; RI: Ramus Intermedius; RCA, right coronary artery; D1/D2, diagonal branches; OM1/OM2, obtuse marginal branches; LVP, left ventricular posterior branch; PDA, posterior descending artery.

## Extended Table 11. Thrombus Classification by Coronary Segment

| Segment | n (%) | AUROC (95% CI) | AUPRC (95% CI) | Sens (95% CI) | Spec (95% CI) | Youden threshold |
|---|---|---|---|---|---|---|
| Left Main | 4/4828 (0.08%) | 0.895 (0.719-0.972) | 0.010 (0.002-0.025) | 1.00 (1.00-1.00) | 0.71 (0.70-0.72) | - |
| Proximal LAD | 28/4828 (0.6%) | 0.946 (0.901-0.981) | 0.401 (0.223-0.574) | 0.89 (0.77-1.00) | 0.89 (0.88-0.90) | 0.0086 |
| Mid LAD | 43/4828 (0.9%) | 0.907 (0.869-0.942) | 0.214 (0.105-0.337) | 0.88 (0.78-0.97) | 0.76 (0.75-0.77) | 0.0038 |
| Distal LAD | 4/4828 (0.08%) | 0.588 (0.272-0.907) | 0.008 (0.000-0.035) | 1.00 (1.00-1.00) | 0.26 (0.25-0.28) | - |
| D1 | 2/4828 (0.04%) | 0.904 (0.806-0.996) | 0.033 (0.001-0.120) | 1.00 (1.00-1.00) | 0.81 (0.80-0.82) | - |

| Segment | n (%) | AUROC (95% CI) | AUPRC (95% CI) | Sens (95% CI) | Spec (95% CI) | Youden threshold |
|---|---|---|---|---|---|---|
| D2 | 2/4828 (0.04%) | 0.603 (0.325-0.872) | 0.002 (0.000-0.005) | 0.51 (0.00-1.00) | 0.87 (0.86-0.88) | - |
| Rx | 4/4828 (0.08%) | 0.607 (0.240-0.820) | 0.002 (0.001-0.005) | 0.75 (0.25-1.00) | 0.75 (0.74-0.76) | - |
| Proximal LCX | 8/4828 (0.2%) | 0.682 (0.575-0.797) | 0.004 (0.001-0.007) | 1.00 (1.00-1.00) | 0.43 (0.42-0.45) | 0.0003 |
| Mid LCX | 11/4828 (0.2%) | 0.778 (0.605-0.927) | 0.027 (0.005-0.065) | 0.73 (0.43-1.00) | 0.76 (0.74-0.77) | 0.0028 |
| Distal LCX | 3/4828 (0.06%) | 0.776 (0.370-0.997) | 0.038 (0.000-0.151) | 0.67 (0.00-1.00) | 0.95 (0.94-0.95) | - |
| LVP | 0/4828 (0.0%) | N/A | N/A | N/A | N/A | - |
| OM1 | 4/4828 (0.08%) | 0.803 (0.580-0.965) | 0.006 (0.000-0.018) | 1.00 (1.00-1.00) | 0.58 (0.56-0.59) | - |
| OM2 | 2/4828 (0.04%) | 0.533 (0.174-0.911) | 0.002 (0.000-0.007) | 0.49 (0.00-1.00) | 0.91 (0.90-0.91) | - |
| Proximal RCA | 43/4828 (0.9%) | 0.935 (0.881-0.976) | 0.441 (0.272-0.589) | 0.93 (0.85-1.00) | 0.85 (0.84-0.86) | 0.0109 |
| Mid RCA | 37/4828 (0.8%) | 0.925 (0.870-0.966) | 0.300 (0.153-0.447) | 0.87 (0.75-0.97) | 0.88 (0.87-0.89) | 0.0126 |
| Distal RCA | 16/4828 (0.3%) | 0.964 (0.929-0.991) | 0.237 (0.079-0.442) | 1.00 (1.00-1.00) | 0.81 (0.80-0.82) | 0.0058 |
| PDA | 2/4828 (0.04%) | 0.098 (0.041-0.159) | 0.000 (0.000-0.001) | 1.00 (1.00-1.00) | 0.05 (0.04-0.05) | - |
| Posterolateral | 6/4828 (0.1%) | 0.832 (0.721-0.938) | 0.016 (0.001-0.058) | 1.00 (1.00-1.00) | 0.71 (0.70-0.73) | 0.0017 |

**Legend.** LVP excluded due to no thrombus cases. Values in parentheses represent 95% confidence intervals.

**Abbreviations:** n, thrombus cases/total segments; AUROC, area under the receiver operating characteristic curve; AUPRC, area under the precision-recall curve; Sens, sensitivity; Spec, specificity; LAD, left anterior descending; LCX, left circumflex; RI: Ramus Intermedius; RCA, right coronary artery; D1/D2, diagonal branches; OM1/OM2, obtuse marginal branches; LVP, left ventricular posterior branch; PDA, posterior descending artery.

## Extended Table 12. Chronic Total Occlusion Classification by Coronary Segment

| Segment | n (%) | AUROC (95% CI) | AUPRC (95% CI) | Sens (95% CI) | Spec (95% CI) |
|---|---|---|---|---|---|
| Left Main | 3/4828 (0.06%) | 0.966 (0.941-0.992) | 0.021 (0.004-0.058) | 1.00 (1.00-1.00) | 0.94 (0.94-0.95) |
| Proximal LAD | 33/4828 (0.7%) | 0.930 (0.886-0.967) | 0.171 (0.081-0.283) | 0.91 (0.80-1.00) | 0.82 (0.81-0.83) |
| Mid LAD | 70/4828 (1.4%) | 0.938 (0.917-0.957) | 0.205 (0.139-0.282) | 0.93 (0.86-0.98) | 0.83 (0.82-0.84) |
| Distal LAD | 11/4828 (0.2%) | 0.808 (0.672-0.927) | 0.013 (0.004-0.028) | 0.82 (0.55-1.00) | 0.72 (0.71-0.73) |
| D1 | 20/4828 (0.4%) | 0.779 (0.691-0.860) | 0.016 (0.007-0.030) | 0.95 (0.83-1.00) | 0.55 (0.53-0.56) |
| D2 | 8/4828 (0.2%) | 0.728 (0.596-0.861) | 0.006 (0.001-0.015) | 1.00 (1.00-1.00) | 0.47 (0.46-0.48) |
| RI | 5/4828 (0.1%) | 0.860 (0.758-0.944) | 0.007 (0.002-0.016) | 1.00 (1.00-1.00) | 0.71 (0.70-0.73) |

| Segment | n (%) | AUROC (95% CI) | AUPRC (95% CI) | Sens (95% CI) | Spec (95% CI) |
| --- | --- | --- | --- | --- | --- |
| Proximal LCX | 46/4828 (1.0%) | 0.896 (0.854-0.933) | 0.122 (0.065-0.204) | 0.87 (0.77-0.96) | 0.79 (0.78-0.80) |
| Mid LCX | 37/4828 (0.8%) | 0.862 (0.815-0.903) | 0.064 (0.027-0.127) | 0.75 (0.61-0.89) | 0.84 (0.83-0.85) |
| Distal LCX | 7/4828 (0.1%) | 0.881 (0.783-0.960) | 0.018 (0.003-0.055) | 1.00 (1.00-1.00) | 0.66 (0.65-0.68) |
| LVP | 4/4828 (0.08%) | 0.857 (0.573-0.994) | 0.022 (0.001-0.085) | 0.76 (0.00-1.00) | 0.90 (0.89-0.91) |
| OM1 | 12/4828 (0.2%) | 0.864 (0.774-0.935) | 0.029 (0.006-0.079) | 1.00 (1.00-1.00) | 0.62 (0.60-0.63) |
| OM2 | 21/4828 (0.4%) | 0.761 (0.674-0.842) | 0.014 (0.007-0.026) | 0.67 (0.47-0.86) | 0.81 (0.80-0.82) |
| Proximal RCA | 134/4828 (2.8%) | 0.958 (0.947-0.969) | 0.374 (0.300-0.459) | 0.96 (0.92-0.99) | 0.87 (0.86-0.88) |
| Mid RCA | 89/4828 (1.8%) | 0.949 (0.933-0.962) | 0.250 (0.179-0.323) | 0.93 (0.88-0.98) | 0.85 (0.84-0.86) |
| Distal RCA | 23/4828 (0.5%) | 0.927 (0.895-0.955) | 0.082 (0.030-0.161) | 1.00 (1.00-1.00) | 0.79 (0.78-0.81) |
| PDA | 15/4828 (0.3%) | 0.791 (0.658-0.910) | 0.022 (0.007-0.048) | 0.80 (0.58-1.00) | 0.74 (0.73-0.75) |
| Posterolateral | 18/4828 (0.4%) | 0.888 (0.837-0.931) | 0.024 (0.011-0.044) | 0.89 (0.71-1.00) | 0.81 (0.80-0.82) |

**Abbreviations:** n, chronic total occlusion cases/total segments; AUROC, area under the receiver operating characteristic curve; AUPRC, area under the precision-recall curve; Sens, sensitivity; Spec, specificity; LAD, left anterior descending; LCX, left circumflex; CTO, Chronic Total Occlsuion, RI: Ramus Intermedius; RCA, right coronary artery; D1/D2, diagonal branches;

OM1/OM2, obtuse marginal branches; LVP, left ventricular posterior branch; PDA, posterior descending artery.

**Legend:** Values in parentheses represent 95% confidence intervals.

**Extended Table 13. MViT-Kinetics Baseline Linear Probing Performance by Coronary Territory With 95% Confidence Intervals (N=4,828 studies)**

| Task | Metric | Global (All) | LCA | RCA |
|---|---|---|---|---|
| **Stenosis (Regression)** | n (segments) | 86,904 | 62,764 | 24,140 |
| | MAE | 10.4 (10.3-10.5) | 9.9 (9.7-10.0) | 11.9 (11.6-12.1) |
| | Pearson | 0.407 (0.398-0.416) | 0.374 (0.366-0.384) | 0.461 (0.449-0.473) |
| **Stenosis ≥70%** | n (abnormal/total) | 7,200/86,904 (8.3%) | 4,939/62,764 (7.9%) | 2,261/24,140 (9.4%) |
| | AUROC | 0.769 (0.763-0.775) | 0.771 (0.765-0.778) | 0.769 (0.759-0.778) |
| | AUPRC | 0.286 (0.272-0.296) | 0.266 (0.255-0.277) | 0.331 (0.315-0.352) |
| | Sens | 0.762 (0.743-0.789) | 0.748 (0.718-0.782) | 0.594 (0.555-0.611) |
| | Spec | 0.633 (0.608-0.648) | 0.660 (0.634-0.691) | 0.819 (0.811-0.854) |
| **Calcification** | n (abnormal/total) | 6,886/86,904 (7.9%) | 4,740/62,764 (7.6%) | 2,146/24,140 (8.9%) |
| | AUROC | 0.767 (0.762-0.773) | 0.767 (0.761-0.775) | 0.772 (0.761-0.782) |

| Task | Metric | Global (All) | LCA | RCA |
|---|---|---|---|---|
| **CTO** | AUPRC | 0.263 (0.252-0.273) | 0.243 (0.232-0.255) | 0.309 (0.292-0.332) |
| | Sens | 0.765 (0.747-0.793) | 0.759 (0.724-0.781) | 0.598 (0.559-0.617) |
| | Spec | 0.632 (0.603-0.648) | 0.647 (0.630-0.678) | 0.817 (0.809-0.852) |
| | n (abnormal/total) | 556/86,904 (0.64%) | 277/62,764 (0.44%) | 279/24,140 (1.2%) |
| | AUROC | 0.801 (0.787-0.819) | 0.771 (0.745-0.799) | 0.818 (0.791-0.843) |
| **Thrombus** | AUPRC | 0.046 (0.039-0.058) | 0.017 (0.013-0.023) | 0.090 (0.067-0.118) |
| | Sens | 0.597 (0.560-0.788) | 0.773 (0.719-0.874) | 0.677 (0.642-0.779) |
| | Spec | 0.858 (0.666-0.883) | 0.645 (0.538-0.684) | 0.849 (0.785-0.864) |
| | n (abnormal/total) | 219/86,904 (0.25%) | 115/62,764 (0.18%) | 104/24,140 (0.43%) |
| | AUROC | 0.769 (0.742-0.801) | 0.767 (0.719-0.809) | 0.744 (0.693-0.781) |
| | AUPRC | 0.007 (0.006-0.009) | 0.005 (0.004-0.007) | 0.010 (0.007-0.013) |
| | Sens | 0.721 (0.636-0.794) | 0.687 (0.607-0.800) | 0.692 (0.605-0.915) |
| | Spec | 0.750 (0.696-0.811) | 0.797 (0.676-0.842) | 0.725 (0.523-0.777) |

**Abbreviations:** LCA, left coronary artery; RCA, right coronary artery; MAE, mean absolute error (%); AUROC, area under the receiver operating characteristic curve; AUPRC, area under the precision-recall curve; Sens, sensitivity; Spec, specificity; CTO, chronic total occlusion. Global metrics represent micro-averaged performance across all 18 coronary segments. 95% confidence intervals computed via 100 bootstrap iterations.

**Extended Table 14. Segment-level for calcifications performance of MViTv2-S pretrained on Kinetics-400 for coronary angiography analysis (N=4,828 studies)**

| Region | AUROC (95% CI) | AUPRC (95% CI) | Sens (95% CI) | Spec (95% CI) | r (95% CI) | MAE (95% CI) |
|---|---|---|---|---|---|---|
| **Global** | 0.769 (0.763-0.775) | 0.286 (0.272-0.296) | 0.76 (0.74-0.79) | 0.63 (0.61-0.65) | 0.41 (0.40-0.42) | 10.4 (10.3-10.5) |
| **LCA Territory** | 0.771 (0.765-0.778) | 0.266 (0.255-0.277) | 0.75 (0.72-0.78) | 0.66 (0.63-0.69) | 0.37 (0.37-0.38) | 9.9 (9.7-10.0) |
| Left Main | 0.820 (0.797-0.843) | 0.240 (0.194-0.284) | 0.78 (0.75-0.88) | 0.71 (0.59-0.74) | 0.41 (0.37-0.44) | 8.5 (8.0-9.0) |
| Proximal LAD | 0.751 (0.739-0.767) | 0.382 (0.352-0.410) | 0.73 (0.62-0.79) | 0.64 (0.57-0.76) | 0.39 (0.37-0.41) | 19.6 (19.0-20.1) |
| Mid LAD | 0.740 (0.725-0.755) | 0.420 (0.395-0.453) | 0.75 (0.65-0.84) | 0.61 (0.54-0.72) | 0.39 (0.37-0.41) | 22.9 (22.4-23.5) |
| Distal LAD | 0.612 (0.567-0.661) | 0.058 (0.046-0.090) | 0.46 (0.35-0.83) | 0.72 (0.38-0.86) | 0.10 (0.07-0.14) | 4.3 (4.0-4.7) |
| D1 | 0.685 (0.669-0.709) | 0.194 (0.176-0.215) | 0.70 (0.63-0.82) | 0.60 (0.49-0.67) | 0.23 (0.20-0.25) | 14.5 (13.9-15.3) |
| D2 | 0.630 (0.598-0.658) | 0.094 (0.081-0.112) | 0.57 (0.48-0.79) | 0.65 (0.43-0.74) | 0.12 (0.09-0.15) | 7.6 (7.0-8.1) |
| Proximal LCx | 0.809 (0.791-0.826) | 0.346 (0.300-0.389) | 0.79 (0.75-0.83) | 0.71 (0.69-0.75) | 0.42 (0.39-0.46) | 15.9 (15.4-16.5) |

| Region | AUROC (95% CI) | AUPRC (95% CI) | Sens (95% CI) | Spec (95% CI) | r (95% CI) | MAE (95% CI) |
| --- | --- | --- | --- | --- | --- | --- |
| Mid LCx | 0.705 (0.678-0.731) | 0.147 (0.124-0.173) | 0.71 (0.60-0.83) | 0.61 (0.47-0.70) | 0.21 (0.18-0.23) | 9.8 (9.2-10.4) |
| Distal LCx | 0.573 (0.497-0.639) | 0.031 (0.021-0.048) | 0.42 (0.29-0.62) | 0.77 (0.56-0.87) | 0.05 (0.01-0.08) | 2.5 (2.1-2.8) |
| OM1 | 0.741 (0.715-0.760) | 0.155 (0.125-0.183) | 0.81 (0.66-0.87) | 0.54 (0.49-0.70) | 0.24 (0.21-0.27) | 8.4 (7.8-8.9) |
| OM2 | 0.712 (0.692-0.737) | 0.142 (0.126-0.166) | 0.85 (0.66-0.93) | 0.46 (0.38-0.67) | 0.22 (0.19-0.24) | 8.9 (8.3-9.4) |
| RI | 0.670 (0.626-0.710) | 0.063 (0.048-0.081) | 0.81 (0.51-0.88) | 0.46 (0.36-0.78) | 0.10 (0.07-0.13) | 4.6 (4.0-5.1) |
| LVP | 0.588 (0.449-0.711) | 0.007 (0.005-0.013) | 0.57 (0.25-1.00) | 0.66 (0.14-0.92) | 0.02 (-0.02-0.04) | 0.5 (0.4-0.7) |
| **RCA Territory** | 0.769 (0.759-0.778) | 0.331 (0.315-0.352) | 0.59 (0.56-0.61) | 0.82 (0.81-0.85) | 0.46 (0.45-0.47) | 11.9 (11.6-12.1) |
| Proximal RCA | 0.799 (0.786-0.815) | 0.445 (0.419-0.482) | 0.83 (0.75-0.85) | 0.62 (0.62-0.70) | 0.51 (0.49-0.53) | 19.3 (18.8-19.7) |
| Mid RCA | 0.801 (0.787-0.818) | 0.415 (0.389-0.452) | 0.77 (0.72-0.86) | 0.71 (0.64-0.76) | 0.45 (0.43-0.47) | 18.2 (17.6-18.6) |
| Distal RCA | 0.701 (0.678-0.725) | 0.149 (0.129-0.181) | 0.71 (0.58-0.82) | 0.61 (0.52-0.74) | 0.27 (0.24-0.30) | 10.3 (9.7-11.0) |
| PDA | 0.711 (0.681-0.735) | 0.109 (0.092-0.133) | 0.76 (0.62-0.81) | 0.58 (0.55-0.72) | 0.25 (0.22-0.27) | 6.9 (6.4-7.5) |
| Posterolateral | 0.667 (0.621-0.716) | 0.073 (0.060-0.098) | 0.63 (0.49-0.74) | 0.67 (0.61-0.79) | 0.16 (0.13-0.20) | 4.7 (4.3-5.2) |

**Abbreviations:** LCA, left coronary artery territory; RCA, right coronary artery territory; RI: Ramus Intermedius; LM, left main; LAD, left anterior descending; LCx, left circumflex; D1/D2, first/second diagonal; OM1/OM2, first/second obtuse marginal; LVP, left ventricular posterior branch; PDA, posterior descending artery; AUROC, area under the receiver operating characteristic curve; AUPRC, area under the precision-recall curve; Sens, sensitivity; Spec, specificity.

**Extended Table 15. Segment-level for thrombus performance of MViTv2-S pretrained on Kinetics-400 for coronary angiography analysis (N=4,238 studies)**

| Segment | AUROC (95% CI) | AUPRC (95% CI) | Sensitivity (95% CI) | Specificity (95% CI) |
| --- | --- | --- | --- | --- |
| Left Main | 0.674 (0.388-0.995) | 0.016 (0.001-0.113) | 0.75 (0.50-1.00) | 0.63 (0.35-1.00) |
| Proximal LAD | 0.583 (0.491-0.664) | 0.008 (0.005-0.014) | 0.71 (0.30-1.00) | 0.47 (0.15-0.92) |
| Mid LAD | 0.473 (0.410-0.526) | 0.008 (0.006-0.010) | 0.95 (0.75-1.00) | 0.19 (0.13-0.39) |
| Distal LAD | 0.490 (0.057-0.860) | 0.001 (0.000-0.004) | 0.50 (0.28-1.00) | 0.75 (0.02-0.87) |
| D1 | 0.394 (0.067-0.726) | 0.001 (0.000-0.003) | 0.50 (0.50-1.00) | 0.71 (0.07-0.73) |
| D2 | 0.453 (0.376-0.530) | 0.001 (0.000-0.001) | 1.00 (1.00-1.00) | 0.39 (0.37-0.53) |
| Proximal LCx | 0.497 (0.326-0.784) | 0.002 (0.001-0.005) | 0.50 (0.24-1.00) | 0.66 (0.06-0.93) |
| Mid LCx | 0.719 (0.541-0.876) | 0.007 (0.002-0.024) | 0.64 (0.46-1.00) | 0.75 (0.34-0.94) |

| Segment | AUROC (95% CI) | AUPRC (95% CI) | Sensitivity (95% CI) | Specificity (95% CI) |
|---|---|---|---|---|
| Distal LCx | 0.404 (0.108-0.697) | 0.001 (0.000-0.002) | 1.00 (0.50-1.00) | 0.10 (0.10-0.70) |
| OM1 | 0.437 (0.295-0.566) | 0.001 (0.000-0.002) | 1.00 (1.00-1.00) | 0.29 (0.28-0.54) |
| OM2 | 0.758 (0.529-0.979) | 0.005 (0.000-0.023) | 1.00 (0.67-1.00) | 0.54 (0.53-0.98) |
| RI | 0.488 (0.099-0.826) | 0.001 (0.000-0.005) | 0.50 (0.33-1.00) | 0.70 (0.10-0.88) |
| LVP | N/A | N/A | N/A | N/A |
| Proximal RCA | 0.593 (0.526-0.654) | 0.011 (0.007-0.014) | 0.70 (0.55-0.99) | 0.52 (0.22-0.64) |
| Mid RCA | 0.654 (0.573-0.738) | 0.012 (0.008-0.020) | 0.76 (0.48-0.93) | 0.54 (0.30-0.74) |
| Distal RCA | 0.653 (0.546-0.758) | 0.006 (0.003-0.012) | 0.69 (0.55-1.00) | 0.69 (0.43-0.72) |
| PDA | 0.309 (0.052-0.572) | 0.000 (0.000-0.001) | 0.50 (0.50-1.00) | 0.56 (0.05-0.58) |
| Posterolateral | 0.609 (0.276-0.819) | 0.002 (0.001-0.007) | 0.67 (0.41-1.00) | 0.72 (0.12-0.88) |

**Abbreviations:** LCA, left coronary artery territory; RCA, right coronary artery territory; RI: Ramus Intermedius; LM, left main; LAD, left anterior descending; LCx, left circumflex; D1/D2, first/second diagonal; OM1/OM2, first/second obtuse marginal; LVP, left ventricular posterior branch; PDA, posterior descending artery;AUROC, area under the receiver operating characteristic curve; AUPRC, area under the precision-recall curve; Sens, sensitivity; Spec, specificity.

**Extended Table 16. Classification Performance for LVEF Prediction (MHI Test Set, n = 2,530)**

| Metric | DeepCORO-CLIP (95% CI) | CathEF (X3D) (95% CI) |
|---|---|---|
| **LVEF < 40%** | | |
| n (pos/total) | 432/2,530 (17.1%) | 432/2,530 (17.1%) |
| AUROC | 0.861 (0.841–0.879) | 0.828 (0.806–0.847) |
| Sensitivity | 0.78 (0.69–0.84) | 0.77 (0.69–0.85) |
| Specificity | 0.79 (0.72–0.87) | 0.75 (0.65–0.82) |
| PPV | 0.43 (0.37–0.53) | 0.38 (0.33–0.45) |
| NPV | 0.95 (0.93–0.96) | 0.94 (0.92–0.96) |
| **LVEF < 50%** | | |
| n (pos/total) | 867/2,530 (34.3%) | 867/2,530 (34.3%) |
| AUROC | 0.815 (0.797–0.832) | 0.763 (0.743–0.780) |
| Sensitivity | 0.71 (0.62–0.80) | 0.68 (0.58–0.78) |
| Specificity | 0.77 (0.68–0.86) | 0.72 (0.62–0.81) |
| PPV | 0.61 (0.56–0.70) | 0.56 (0.51–0.61) |
| NPV | 0.84 (0.81–0.87) | 0.81 (0.79–0.85) |

**Abbreviations:** LVEF left ventricular ejection fraction; MHI, Montreal Heart Institute; EF, ejection fraction; AUROC, area under the receiver operating characteristic curve; PPV, positive predictive value; NPV, negative predictive value; CI, confidence interval.

**Extended Table 17. Regression Performance for LVEF Prediction (MHI Test Set, n = 2,530)**

| Metric | DeepCORO-CLIP (95% CI) | CathEF (X3D; 95% CI) | Delta |
|---|---|---|---|
| MAE (EF %) | 7.30 (7.03–7.54) | 8.78 (8.55–9.02) | −1.48 |
| RMSE (EF %) | 9.65 (9.32–9.96) | 10.70 (10.44–10.95) | −1.05 |
| Pearson r | 0.65 (0.62–0.68) | 0.50 (0.47–0.52) | +0.15 |

**Abbreviation:** MAE, mean absolute error; RMSE, root mean

**Extended Table 18. DeepCORO-CLIP vs Core lab adjudication vs Clinical Report compared to previous state-of-the-art (DeepCORO)**

| Metric | LCA (n = 476 lesions, 332 studies) | | RCA (n = 489 lesions, 330 studies) | | TOTAL (n = 965 lesions, 662 studies) | |
|---|---|---|---|---|---|---|
| | vs Adjud. | vs Clinical Report | vs Adjud. | vs Clinical Report | vs Adjud. | vs Clinical Report |
| **VIDEO LEVEL** | | | | | | |
| **DeepCORO-CLIP** | | | | | | |
| MAE (%) | 16.71 (14.82-18.58) | 20.65 (18.82–22.75) | 15.13 (13.28-17.10) | 19.74 (17.88-21.85) | 15.89 (14.57-17.23) | 20.18 (18.67-21.65) |
| Corr r | 0.56 (0.46-0.65) | 0.51 (0.42-0.59) | 0.57 (0.48-0.66) | 0.52 (0.44-0.60) | 0.56 (0.48-0.61) | 0.51 (0.45-0.57) |
| **STUDY LEVEL** | | | | | | |
| **DeepCORO-CLIP** | | | | | | |
| MAE (%) | 14.36 (12.60-16.30) | 19.68 (17.95-21.79) | 12.75 (10.76-15.04) | 18.27 (16.83-20.74) | 13.56 (12.51-15.17) | 18.97 (17.55-20.52) |
| Corr r | 0.65 (0.58-0.72) | 0.57 (0.49-0.65) | 0.63 (0.50-0.71) | 0.55 (0.45-0.62) | 0.64 (0.58-0.70) | 0.56 (0.49-0.62) |
| **DeepCORO** | | | | | | |
| MAE (%) | 19.76 (18.35-20.91) | 20.58 (19.23-22.13) | 18.44 (17.09-20.05) | 17.29 (15.66-18.82) | 19.09 (18.16-20.13) | 18.92 (17.88-19.95) |
| Corr r | 0.66 (0.60-0.70) | 0.46 (0.38-0.53) | 0.70 (0.65-0.75) | 0.59 (0.52-0.66) | 0.68 (0.64-0.72) | 0.52 (0.47-0.57) |

**Abbreviations:** LCA, Left Coronary Artery; RCA, Right Coronary Artery; Adjud., Adjudication (expert human annotation); Report, clinical report reference; Mean, average of Adjudication and Report; MAE, Mean Absolute Error; Corr r, Pearson correlation coefficient. Values in parentheses are 95% confidence intervals from bootstrap resampling (n=1000).

**Extended Table 19. UCSF External Validation Results - Stenosis ≥70 (n=4,249 studies)**

| Artery Segment | n | MAE (%) (95% CI) | Pearson r (95% CI) | AUROC (95% CI) | AUPRC (95% CI) | Sens (95% CI) | Spec (95% CI) |
|---|---|---|---|---|---|---|---|
| **Global** | 1794/4249 (42.2%) | 12.1 (11.9-12.3) | 0.45 (0.44-0.46) | 0.89 (0.88-0.89) | 0.40 (0.39-0.42) | 0.84 (0.82-0.88) | 0.77 (0.74-0.80) |
| **Left Coronary Artery** | 1554/4249 (36.6%) | 13.0 (12.8-13.2) | 0.43 (0.42-0.44) | 0.87 (0.87-0.88) | 0.37 (0.35-0.38) | 0.84 (0.81-0.87) | 0.75 (0.71-0.79) |
| Left Main | 94/4249 (2.2%) | 8.7 (8.2-9.1) | 0.53 (0.49-0.57) | 0.95 (0.94-0.96) | 0.33 (0.24-0.43) | 0.87 (0.85-0.98) | 0.92 (0.80-0.93) |
| Proximal LAD | 573/4249 (13.5%) | 18.8 (18.2-19.4) | 0.59 (0.56-0.61) | 0.89 (0.88-0.90) | 0.59 (0.54-0.63) | 0.86 (0.79-0.89) | 0.78 (0.75-0.84) |
| Mid LAD | 586/4249 (13.8%) | 20.9 (20.2-21.5) | 0.55 (0.52-0.58) | 0.85 (0.84-0.86) | 0.47 (0.43-0.51) | 0.83 (0.73-0.88) | 0.71 (0.66-0.82) |
| Distal LAD | 181/4249 (4.3%) | 12.1 (11.4-12.8) | 0.20 (0.17-0.23) | 0.80 (0.77-0.83) | 0.14 (0.11-0.18) | 0.80 (0.72-0.91) | 0.70 (0.59-0.77) |
| D1 | 317/4249 (7.5%) | 16.0 (15.1-16.8) | 0.37 (0.34-0.40) | 0.82 (0.80-0.84) | 0.22 (0.19-0.26) | 0.87 (0.81-0.94) | 0.66 (0.58-0.72) |

| Artery Segment | n | MAE (%) (95% CI) | Pearson r (95% CI) | AUROC (95% CI) | AUPRC (95% CI) | Sens (95% CI) | Spec (95% CI) |
|---|---|---|---|---|---|---|---|
| D2 | 146/4249 (3.4%) | 7.0 (6.5-7.6) | 0.17 (0.13-0.21) | 0.81 (0.79-0.84) | 0.10 (0.08-0.14) | 0.88 (0.77-0.96) | 0.63 (0.53-0.77) |
| Proximal LCX | 317/4249 (7.5%) | 15.0 (14.5-15.5) | 0.58 (0.55-0.61) | 0.92 (0.90-0.93) | 0.49 (0.43-0.55) | 0.85 (0.82-0.96) | 0.83 (0.72-0.86) |
| Mid LCX | 230/4249 (5.4%) | 13.8 (13.1-14.6) | 0.38 (0.35-0.42) | 0.86 (0.84-0.88) | 0.26 (0.22-0.33) | 0.90 (0.82-0.93) | 0.73 (0.72-0.80) |
| Distal LCX | 98/4249 (2.3%) | 7.7 (7.1-8.2) | 0.15 (0.11-0.19) | 0.86 (0.84-0.89) | 0.13 (0.09-0.20) | 0.93 (0.78-0.99) | 0.67 (0.61-0.82) |
| OM1 | 280/4249 (6.6%) | 14.3 (13.5-15.1) | 0.33 (0.30-0.37) | 0.83 (0.81-0.85) | 0.22 (0.18-0.26) | 0.81 (0.78-0.92) | 0.74 (0.62-0.76) |
| OM2 | 193/4249 (4.5%) | 9.0 (8.3-9.6) | 0.28 (0.23-0.31) | 0.82 (0.80-0.84) | 0.17 (0.13-0.21) | 0.89 (0.75-0.93) | 0.63 (0.59-0.76) |
| **Right Coronary Artery** | 993/4249 (23.4%) | 10.4 (10.2-10.7) | 0.50 (0.49-0.52) | 0.92 (0.91-0.92) | 0.47 (0.44-0.50) | 0.88 (0.84-0.90) | 0.79 (0.77-0.84) |
| Proximal RCA | 422/4249 (9.9%) | 16.7 (16.1-17.2) | 0.63 (0.60-0.65) | 0.92 (0.90-0.93) | 0.59 (0.54-0.64) | 0.90 (0.83-0.94) | 0.79 (0.76-0.86) |

| Artery Segment | n | MAE (%) (95% CI) | Pearson r (95% CI) | AUROC (95% CI) | AUPRC (95% CI) | Sens (95% CI) | Spec (95% CI) |
| --- | --- | --- | --- | --- | --- | --- | --- |
| Mid RCA | 448/4249 (10.5%) | 16.6 (16.0-17.3) | 0.58 (0.55-0.61) | 0.90 (0.89-0.92) | 0.57 (0.52-0.62) | 0.87 (0.78-0.93) | 0.76 (0.72-0.85) |
| Distal RCA | 235/4249 (5.5%) | 13.3 (12.5-14.0) | 0.39 (0.35-0.42) | 0.88 (0.86-0.89) | 0.33 (0.28-0.40) | 0.89 (0.82-0.93) | 0.73 (0.70-0.81) |
| PDA | 166/4249 (3.9%) | 9.2 (8.5-9.8) | 0.37 (0.33-0.41) | 0.86 (0.83-0.88) | 0.23 (0.17-0.29) | 0.83 (0.73-0.89) | 0.73 (0.69-0.83) |
| Posterolateral | 113/4249 (2.7%) | 6.8 (6.3-7.4) | 0.13 (0.09-0.18) | 0.83 (0.80-0.86) | 0.11 (0.08-0.15) | 0.88 (0.78-0.95) | 0.67 (0.58-0.78) |

**Abbreviations:** n, studies with abnormality/total (prevalence %); Sens, sensitivity; Spec, specificity; MAE, mean absolute error; AUROC, area under receiver operating characteristic curve; AUPRC, area under precision-recall curve; CI, confidence interval; LAD, left anterior descending artery; LCX, left circumflex artery; LCA, left coronary artery; RCA, right coronary artery; D1, first diagonal branch; D2, second diagonal branch; OM1, first obtuse marginal branch; OM2, second obtuse marginal branch; PDA, posterior descending artery.

**Legend:** Optimal threshold for sensitivity and specificity per-segment using Youden's J statistic (J = Sens + Spec - 1)